\pdfoutput=1
\documentclass[11pt]{article}

\usepackage[final]{coling}

\usepackage{times}
\usepackage{latexsym}

\usepackage[T1]{fontenc}
\usepackage[utf8]{inputenc}

\usepackage{microtype}

\usepackage{inconsolata}

\usepackage{graphicx}

\usepackage{times}
\usepackage{latexsym}
\usepackage{todonotes}
\usepackage{multirow}
\usepackage{booktabs}
\usepackage{pdflscape}
\usepackage{colortbl}
\usepackage{float}
\usepackage{placeins} % Add this line to include the placeins package
\usepackage{arydshln}
\usepackage{xcolor}
\usepackage{graphicx}
\usepackage{adjustbox}
\usepackage{wrapfig}
\usepackage{hyperref}
\usepackage{subfigure}
\usepackage{longtable}
\usepackage{multirow}
\usepackage[T1]{fontenc}
\usepackage{enumitem}

\newcommand{\ctslogo}{\raisebox{3.4pt}{\includegraphics[scale=0.01]{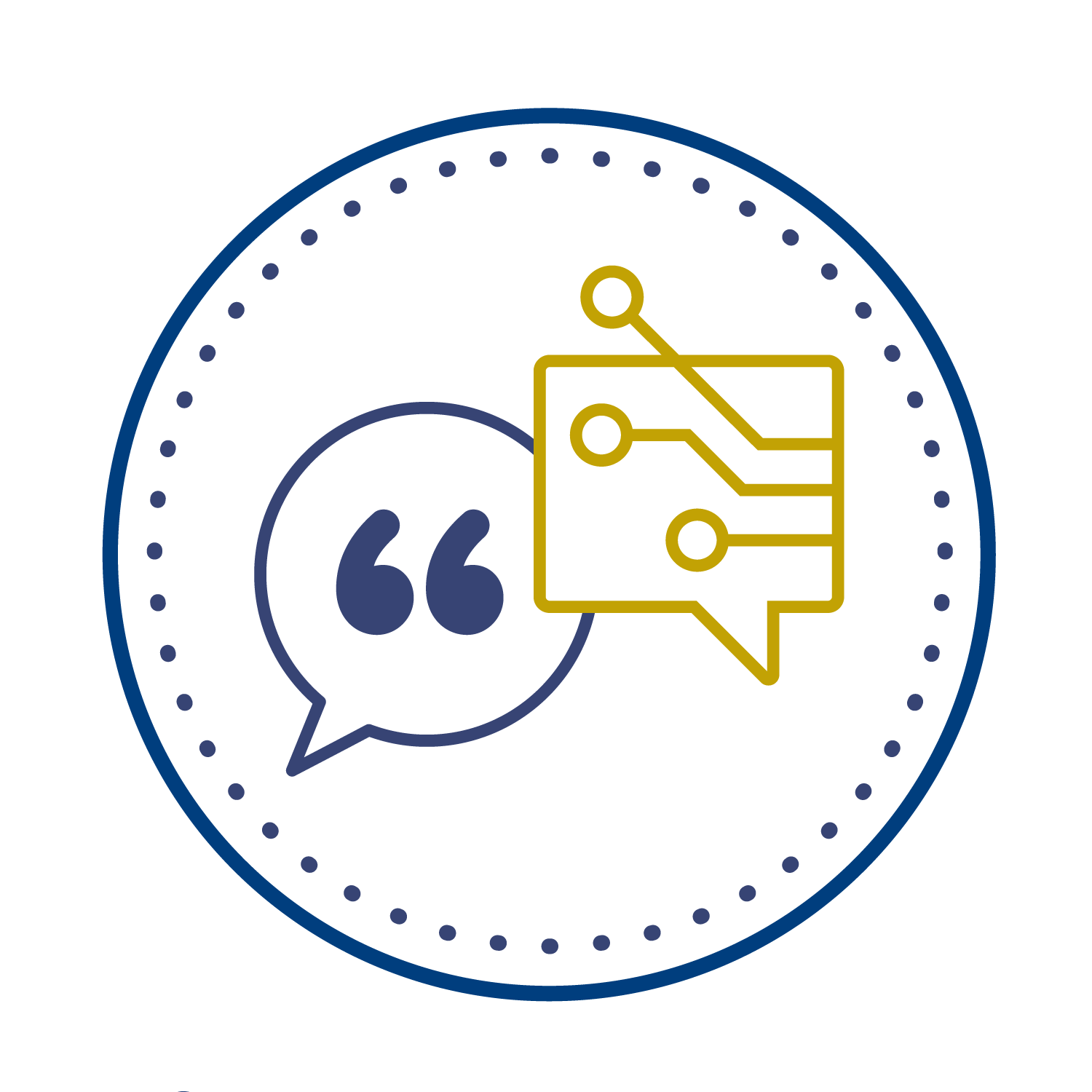}}}
\newcommand{\PAIlogo}{\raisebox{3.4pt}{\includegraphics[scale=0.08]{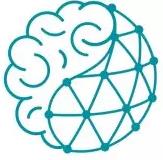}}}

\title{When LLMs Struggle: Reference-less Translation Evaluation \\for Low-resource Languages}

  \author{
     Archchana Sindhujan\PAIlogo, Diptesh Kanojia\PAIlogo,
    \bf{Constantin Or\u{a}san\ctslogo, Shenbin Qian\ctslogo}\\ [.35em]
    \PAIlogo Institute for People-Centred AI and     \ctslogo Centre for Translation Studies, \\ University of Surrey, United Kingdom\\ [.15em]
    \texttt{\{a.sindhujan, d.kanojia, c.orasan, s.qian\}@surrey.ac.uk}\\ 
}
\begin{document}
\maketitle
\begin{abstract}
This paper investigates the \textit{reference-less} evaluation of machine translation for low-resource language pairs, known as quality estimation (QE). Segment-level QE is a challenging cross-lingual language understanding task that provides a quality score  ($0-100$) to the translated output. We comprehensively evaluate large language models (LLMs) in zero/few-shot scenarios and perform instruction fine-tuning using a novel prompt based on annotation guidelines. Our results indicate that prompt-based approaches are outperformed by the encoder-based fine-tuned QE models. Our error analysis reveals tokenization issues, along with errors due to transliteration and named entities, and argues for refinement in LLM \textit{pre-training} for cross-lingual tasks. We release the data, and models trained publicly for further research. 
\end{abstract}

\section{Introduction}

Traditional methods of obtaining references for machine-translated texts are costly, and prone to subjectivity and inconsistency~\cite{rei-etal-2021-references, lo2014xmeant,huynh2008sectra_w}. To address these challenges of evaluating imperfect translations, Quality Estimation (QE) has emerged as a crucial area, enabling the assessment of MT output in the absence of a reference~\cite{zerva-etal-2022-findings}.

Our work investigates segment-level QE~\cite{blain-etal-2023-findings,zerva-etal-2022-findings,fernandes-etal-2023-devil}, which is \textit{conventionally} modelled as a \textit{regression task} and aims to predict a segment-level quality score, also known as the direct assessment (DA) score~\citep{graham-etal-2013-continuous}. Due to the underlying subjectivity in human translation quality evaluation, DA score is computed as a \textit{mean} of three or more human annotations on a scale of {$0-100$}. While large language models (LLMs) claim superlative performance for different natural language processing (NLP) tasks~\cite{devlin2019bert,achiam2023gpt}, evaluation of machine-translated output poses a unique challenge where both \textit{syntactic accuracy} \textit{and cross-lingual semantic match} are relevant, for the prediction of DA scores.

LLMs are applicable for many NLP tasks, including machine translation (MT)~\cite{kocmi-etal-2023-findings, robinson-etal-2023-chatgpt, manakhimova-etal-2023-linguistically} and quality estimation~\cite{kocmi-federmann-2023-large,xu-etal-2023-instructscore,fernandes-etal-2023-devil,huang-etal-2024-lost}. There are significant disparities in the reported performance of LLMs between high- and low-resource languages~\cite{huang2023not,nguyen-etal-2024-democratizing}. LLMs exhibit better performance in evaluating the quality when references are available~\cite{huang-etal-2024-lost}; however, they are challenging to scale due to the cost associated with manual translation. 

This work focuses on the \textit{reference-less} scenario, evaluating the efficacy of LLMs in settings like zero-shot, few-shot/in-context learning (ICL), and instruction fine-tuning with an adapter~\cite{hu2021lora}. We present a novel prompt which utilizes annotation guidelines within prompt instructions and improves task performance. Additionally, we perform experiments for both independent language-pair training (\textit{ILT - training instances from one language pair}), and unified multilingual training (\textit{UMT - training instances from all language pairs}) settings. Our contributions are:

\begin{itemize}[noitemsep, topsep=0pt, leftmargin=*]
 
\item A novel annotation guidelines-based prompt (AG-prompt) which improves zero-shot performance.

\item A comprehensive evaluation for segment-level QE using multiple LLMs, indicating challenges for cross-lingual NLP tasks.

\item Instruction fine-tuned QE model adapters \textit{(4-bit)} for quick deployment.

\item Quantitative and Qualitative analysis indicating critical challenges using LLMs for cross-lingual tasks involving low-resource languages. 
 
\end{itemize}

\section{Background}\label{backgroung}
%- LLMS and QE\\
% Transformer-based methods are leading the way in the field of quality estimation.
Transformer-based approaches which leverage supervised fine-tuning of regression models significantly improved the performance of QE models~\cite{ranasinghe-etal-2020-transquest}. Recently proposed approaches like CometKiwi~\cite{rei-etal-2023-scaling}, Ensemble-CrossQE~\cite{li-etal-2023-hw-tsc} and TransQuest~\cite{ranasinghe-etal-2020-transquest,sindhujan-etal-2023-surreyai} from WMT QE shared tasks~\cite{blain-etal-2023-findings} are based on pre-trained encoder-based language models. However, recent claims have propelled the use of LLMs across various NLP tasks~\cite{zhao2023survey}. Following suit,~\citet{kocmi-federmann-2023-large} introduced the GEMBA prompt-based metric for evaluating translation quality. Their approach focuses on zero-shot prompt-based evaluation, comparing four prompt variants across nine GPT model variants for three high-resource language pairs. The paper discusses experiments with both settings, with and without reference, claiming SoTA performance by including the reference for DA prediction. Our experiments reproduce their prompt in a reference-less setting utilizing only publicly available LLMs and compare prompting strategies by adding relevant context.  

~\citet{huang-etal-2024-lost} examined how LLMs use source and reference information for translation evaluation and they observed that reference information improves accuracy and correlations, while source information shows a negative impact, highlighting limitations in LLMs' cross-lingual semantic matching capability, which is essential for a task such as QE.~\citet{vandan-etal-2023-towards} perform QE by pre-tuning the adapter using a large parallel corpus of English-Indic languages over machine translation task. They fine-tune the model again using supervised QE data and show that pre-tuning the model using MT does not help. 
% Further, they report their performance only on the development set, while we report performance on the test set DA annotations. 
Other approaches to QE such as MQM, include fine-grained error annotation and detailed explanations, which are often not viable for low-resource languages due to lack of annotated data. 

% The MQM framework, which provides fine-grained error annotations with specific error categories and severity levels, is available to only a few low-resource language pairs, such as English to Hindi, Marathi, and Tamil. ~\citet{xu-etal-2023-instructscore} propose fine-grained, explainable evaluation metrics which address the issue of lack of explicit explanations and error identification by providing both a score and a human-readable diagnostic report.~\citet{fernandes-etal-2023-devil} introduced AutoMQM, a technique for evaluating translation quality by prompting LLMs to identify and categorize errors instead of assigning numerical scores. Although this method does not surpass standard automatic metrics, it provides interpretable error spans that align well with human annotations, even without fine-tuning.

 \begin{figure*}[t]
 \centering
 \includegraphics[width=0.9\textwidth, keepaspectratio]{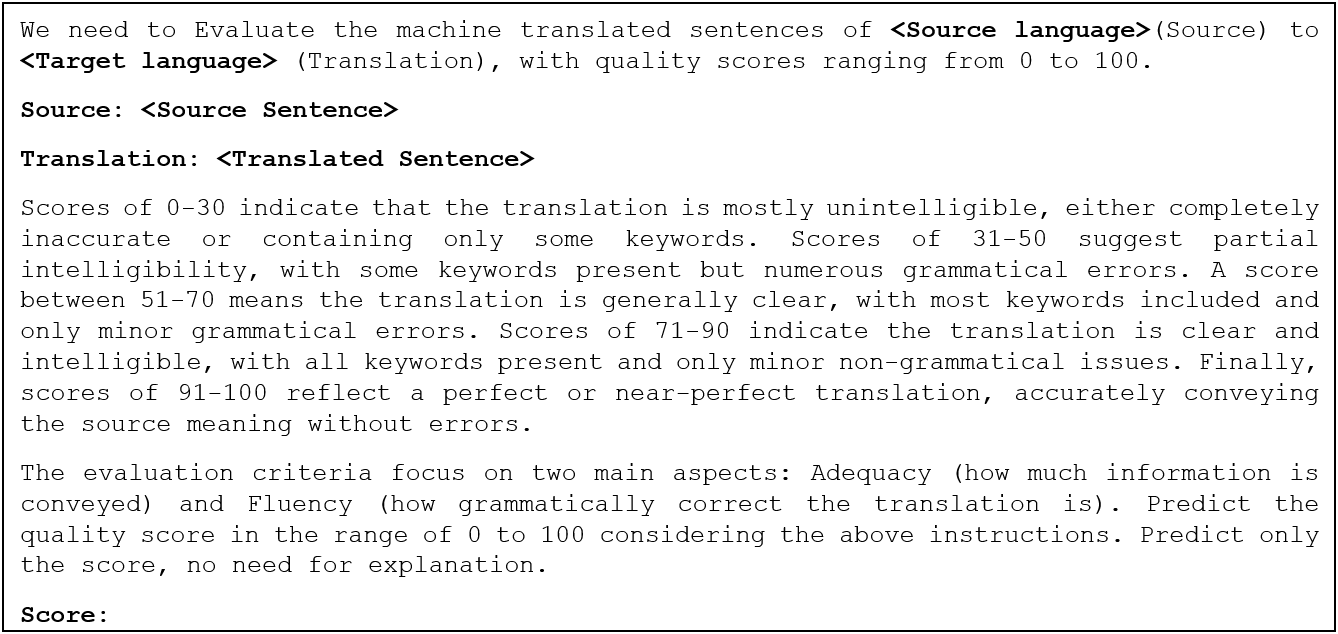}
\caption{The proposed AG prompt which augments scoring instructions within the context.}
\label{fig:zeroshot} % Give a unique label
\end{figure*}

\section{Methodology}\label{sec:methodology}

\subsection{Datasets} \label{subsec:datasets}
% Our study focuses on the low-resource language pairs covered by the WMT QE shared tasks which have human-annotated DA scores\footnote{Mean DA score annotations for Indic-language pair test sets shared upon request}. Low-resource Indic language datasets which have English in the source side of the translation were retrieved from the WMT23 QE shared task~\cite{blain-etal-2023-findings}—English to Gujarati, Hindi, Marathi, Tamil, and Telugu (En-Gu, En-Hi, En-Mr, En-Ta, En-Te). 

%We note that Hindi could be considered a mid-resource language given the parallel corpus, and digitised text available, in the context of machine translation~\cite{nguyen-etal-2024-democratizing}. However, when evaluation and QE are taken into account, Hindi, Marathi, or any other Indic languages, do not have sufficient resources compared to English. 

%Additionally, datasets for other low-resource language pairs which have English in the target side of the translation were sourced from the WMT22 QE shared task~\cite{zerva-etal-2022-findings}—Estonian, Nepali, and Sinhala to English (Et-En, Ne-En, Si-En). A detailed overview of the data splits utilized for our study can be found in Appendix~\ref{app:DA_dataset}. The training splits were used to fine-tune the language models, while the test splits were used for zero-shot, ICL, and to inference from the fine-tuned(FT) models. 

Our study focuses on low-resource language pairs from the WMT QE shared tasks with human-annotated DA scores, including English to Gujarati, Hindi, Marathi, Tamil, and Telugu (En-Gu, En-Hi, En-Mr, En-Ta, En-Te) from WMT23~\cite{blain-etal-2023-findings}. We also include Estonian, Nepali, and Sinhala to English (Et-En, Ne-En, Si-En) language pairs from WMT22~\cite{zerva-etal-2022-findings}. Hindi and Estonian although mid-resource for machine translation~\cite{nguyen-etal-2024-democratizing}, lack sufficient resources for translation evaluation and QE. Training splits were used for fine-tuning, while test splits were used for zero-shot, ICL, and inference experiments (Appendix~\ref{app:DA_dataset}).

\subsection{Prompting Strategies} \label{prompt_variant}

\paragraph{Zero-shot} prompting refers to a model generating outputs for a given input prompt solely based on its pre-trained knowledge and inherent generalization capabilities, without requiring any additional fine-tuning or contextual examples. Existing studies highlight that adding context and reasoning to prompts can significantly enhance LLM's performance in NLP tasks~\cite{zhou-etal-2023-context,chen-etal-2023-many}. However, for the low-resource languages, fine-grained QE data is unavailable. Therefore, we experiment with different prompting strategies: 1) instructing the model to act as a translation evaluator (TE) (Appendix~\ref{app:other_prompts}) and 2) providing additional context from human annotation guidelines (AG). Using the proposed AG prompt (Figure~\ref{fig:zeroshot}), we incorporate reasoning to evaluate translation quality. We compare these strategies with the GEMBA prompt~\cite{kocmi-federmann-2023-large} in the zero-shot setting.

\paragraph{In-context learning} refers to the ability of large language models to perform a task by leveraging examples of the task provided within the input context, without requiring any additional training. We focus our investigation on the AG prompt within the ICL scenario. In this setting, we augmented the AG prompt with example annotations from 5 different  DA score ranges (0-30, 31-50, 51-70, 71-90, 91-100), as detailed in Appendix~\ref{app:icl}. The ICL experiments were divided into three configurations: 3-ICL, 5-ICL, and 7-ICL. In the 5-ICL configuration, we selected one example from each of the five predefined DA score ranges. The 3-ICL configuration excluded examples from the 31-50 and 51-70 ranges. For the 7-ICL configuration, we included one example from each range, plus two additional samples—one from the lowest and one from the highest available score ranges. Through in-context learning experiments, we aim to assess whether incorporating examples of DA annotations can enhance the model's performance. Additionally, by varying the number of examples in each ICL setting, we investigate the impact on the performance of the QE model. 

Furthermore, instruction fine-tuning involves adapting a model using a dataset that includes explicit instructions for specific tasks. In our instruction fine-tuning experiments, we employ the AG prompt to evaluate its effect on model performance.

\subsection{Implementation Details}

\begin{table*}[ht!]
\centering
\tiny
% \small

% \renewcommand{\arraystretch}{1.1}
\resizebox{0.80\textwidth}{!}{
\begin{tabular}{c|rcccc}
\hline
\textbf{LP} & \textbf{Template} & \textbf{Gemma-7B} & \textbf{Llama-2-7B} & \textbf{Llama-2-13B} & \textbf{OC-3.5-7B} \\ 
\hline
\parbox[t]{1mm}{\multirow{6}{*}{\rotatebox[origin=c]{90}{\small En-\textbf{Gu}}}} & \tiny{0-shot-GEMBA}& 0.113 & 0.006 & 0.019 & 0.249\textsuperscript{*}  \\
& \tiny{0-shot-TE} & -0.102\textsuperscript{$\dagger$} & -0.008 & -0.052 & 0.117\textsuperscript{$\dagger$} \\
& \tiny{0-shot-AG} & -0.079 & -0.007 & 0.008 & 0.164\textsuperscript{$\dagger$} \\ \cline{2-6}
& \tiny{3-ICL-AG} & -0.005	& 0.036	& -0.036	& 0.223 \\ 
& \tiny{5-ICL-AG} & 0.023 &	-0.008	& 0.095 &	0.151 \\ 
& \tiny{7-ICL-AG} & 0.071 &	-0.053	& -0.108	& \underline{\textbf{0.260}} \\ 
\hline
\parbox[t]{1mm}{\multirow{6}{*}{\rotatebox[origin=c]{90}{\small En-\textbf{Hi}}}}& \tiny{0-shot-GEMBA} & 0.131 & -0.002 & 0.009 & \textbf{0.254}\textsuperscript{*} \\
& \tiny{0-shot-TE} & -0.050 & -0.072 & 0.056 & 0.134 \\
& \tiny{0-shot-AG} & -0.056 & -0.029 & 0.069 & 0.253 \\  \cline{2-6}
& \tiny{3-ICL-AG} & 0.134 &	-0.114 &	-0.023	& 0.184 \\ 
& \tiny{5-ICL-AG} & 0.075	&-0.022 &	0.035 &	\underline{0.212} \\
& \tiny{7-ICL-AG} & 0.075	&-0.176	& 0.014	& 0.163 \\ 
\hline
\parbox[t]{1mm}{\multirow{6}{*}{\rotatebox[origin=c]{90}{\small En-\textbf{Mr}}}} &  \tiny{0-shot-GEMBA} & 0.135 & 0.053 & 0.115 & 0.183 \\
& \tiny{0-shot-TE} & 0.173 & 0.070 & 0.040 & 0.114 \\
& \tiny{0-shot-AG} & 0.027 & 0.059 & 0.005 & \textbf{0.276}\textsuperscript{*} \\ \cline{2-6}
& \tiny{3-ICL-AG} & 0.202 &	0.120 &	0.095 &	0.218 \\ 
& \tiny{5-ICL-AG} & 0.164\textsuperscript{$\dagger$} &	0.032 &	-0.031 &	0.226 \\
& \tiny{7-ICL-AG} & 0.167 &	0.050 &	0.047 &	\underline{0.251} \\ 
\hline
\parbox[t]{1mm}{\multirow{6}{*}{\rotatebox[origin=c]{90}{\small En-\textbf{Ta}}}} &  \tiny{0-shot-GEMBA} & 0.222 & 0.067 & 0.091 & 0.358 \\
& \tiny{0-shot-TE} & -0.037\textsuperscript{$\dagger$} & 0.012 & 0.016 & 0.178 \\
& \tiny{0-shot-AG} & -0.002 & 0.055 & -0.070 & \textbf{0.363}\textsuperscript{*} \\ \cline{2-6}
& \tiny{3-ICL-AG} & 0.122 &	-0.019 &	0.083 &	\underline{0.337} \\ 
& \tiny{5-ICL-AG} & 0.114 &	0.017 &	0.193 &	0.332 \\
& \tiny{7-ICL-AG} & 0.122 &	-0.096 &	-0.004 &	0.309 \\ 
\hline
\parbox[t]{1mm}{\multirow{6}{*}{\rotatebox[origin=c]{90}{\small En-\textbf{Te}}}} & \tiny{0-shot-GEMBA} & 0.081 & -0.016 & 0.121\textsuperscript{$\dagger$} & 0.145\textsuperscript{*} \\
& \tiny{0-shot-TE} & 0.018 & 0.013 & 0.010 & 0.072 \\
& \tiny{0-shot-AG} & 0.065 & 0.083 & 0.045 & 0.121\textsuperscript{$\dagger$} \\ \cline{2-6}
& \tiny{3-ICL-AG} & 0.092 &	0.027 &	0.015 &	0.152 \\ 
& \tiny{5-ICL-AG} & 0.021 &	0.051 &	0.073 &	0.126 \\
& \tiny{7-ICL-AG} & -0.033 &	0.021 &	-0.028 &	\underline{\textbf{0.196}} \\ 
\hdashline
\parbox[t]{1mm}{\multirow{6}{*}{\rotatebox[origin=c]{90}{\small \textbf{Et}-En}}} &  \tiny{0-shot-GEMBA} & 0.289 & 0.168 & 0.185 & 0.571 \\
& \tiny{0-shot-TE} & 0.086 & 0.100 & 0.146 & 0.455 \\
& \tiny{0-shot-AG} & 0.098 & 0.064 & 0.319 & 0.619\textsuperscript{*} \\ \cline{2-6}
& \tiny{3-ICL-AG} & 0.226 & 0.268 &	-0.058 &	0.613 \\ 
& \tiny{5-ICL-AG} & 0.327 &	0.269 &	0.438 &	\underline{\textbf{0.636}} \\
& \tiny{7-ICL-AG} & 0.306 &	0.033 &	0.169 &	0.616 \\ 
\hline
\parbox[t]{1mm}{\multirow{6}{*}{\rotatebox[origin=c]{90}{\small \textbf{Ne}-En}}} &  \tiny{0-shot-GEMBA} & 0.261 & 0.153 & 0.222 & 0.448 \\
& \tiny{0-shot-TE} & 0.155 & 0.100 & 0.080 & 0.334 \\
& \tiny{0-shot-AG} & 0.130 & 0.144 & 0.303 & 0.487\textsuperscript{*} \\ \cline{2-6}
& \tiny{3-ICL-AG} & 0.273 &	0.149 &	0.340 &	0.457 \\ 
& \tiny{5-ICL-AG} & 0.305 &	0.189 &	0.319 &	0.471 \\
& \tiny{7-ICL-AG} & 0.365 &	-0.040 &	0.259 &	\underline{\textbf{0.491}} \\ 
\hline
\parbox[t]{1mm}{\multirow{6}{*}{\rotatebox[origin=c]{90}{\small \textbf{Si}-En}}} &  \tiny{0-shot-GEMBA} & 0.193 & 0.144 & 0.195 & 0.417 \\
& \tiny{0-shot-TE} & 0.055 & 0.129 & 0.109 & 0.303 \\
& \tiny{0-shot-AG} & 0.042 & 0.069 & 0.238 & 0.441\textsuperscript{*} \\ \cline{2-6}
& \tiny{3-ICL-AG} & 0.306 &	0.146 &	0.018 &	0.470 \\ 
& \tiny{5-ICL-AG} & 0.320\textsuperscript{$\dagger$} &	0.243 &	0.326 &	\underline{\textbf{0.479}} \\
& \tiny{7-ICL-AG} & 0.283 &	-0.017 &	0.223 &	0.477 \\ 
\hline
\end{tabular}
}
\caption{\small Spearman correlation ($\rho$) between the predicted and human-annotated scores for all the experimental settings. Prompt templates: GEMBA, TE, and AG (from section~\ref{prompt_variant}). Bold indicates the overall top score per language pair, asterisks (*) denote top scores in zero-shot settings, and underlined values highlight the best among ICL settings. The (\textsuperscript{$\dagger$}) symbol denotes statistically insignificant results ($p > 0.05$), and  the dashed line separates language pairs with English as target.}
\label{tab:zero_few}
\end{table*}

% For our study, we focus on publicly available LLMs with a parameter count under 13B that have established benchmarks in multilingual performance: Gemma-7B, OpenChat-3.5 and LLaMa-2-(7B/13B) \footnote{\href{https://huggingface.co/google/Gemma-7b}{GEMMA-7B,} \href{https://huggingface.co/OpenChat/OpenChat-3.5-1210}{ OpenChat-3.5,} \href{https://huggingface.co/meta-llama/LLaMA-2-7b-chat-hf}{LLaMa-2-(7B/} \href{https://huggingface.co/meta-llama/LLaMA-2-13b-chat-hf}{13B)}}. 

For our study, we focus on publicly available LLMs with a parameter count under 13B that have established benchmarks in multilingual performance: Gemma-7B\footnote{\href{https://huggingface.co/google/Gemma-7b}{huggingface.co/google/Gemma-7b}}, OpenChat-3.5\footnote{\href{https://huggingface.co/OpenChat/OpenChat-3.5-1210}{ huggingface.co/OpenChat/OpenChat-3.5}}, Llama-2-7B\footnote{\href{https://huggingface.co/meta-llama/LLaMA-2-7b-chat-hf}{huggingface.co/meta-llama/LLaMA-2-7b-chat-hf}}, Llama-2-13B\footnote{\href{https://huggingface.co/meta-llama/LLaMA-2-13b-chat-hf}{huggingface.co/meta-llama/LLaMA-2-13b-chat-hf}}

The OpenChat 7B-parameter model~\citep{Wang2023} (OC-3.5-7B) employs Conditioned-RLFT, a technique that uses a class-conditioned policy to prioritize high-quality responses over sub-optimal ones. The Llama model~\citep{Touvron2023} incorporates supervised fine-tuning (SFT) and reinforcement learning with human feedback (RLHF) to align its outputs with human preferences. Additionally, the Gemma-7B model~\citep{gemmateam2024gemmaopenmodelsbased} utilizes advanced techniques such as Multi-Query Attention, RoPE Embeddings, GeGLU Activations, and RMSNorm to enhance its performance. We chose not to use the latest Llama models (Llama-3 and Llama-3.1) in our experiments, as results from initial zero-shot evaluations showed they did not produce meaningful outputs.

We fine-tune regression models using QE frameworks such as TransQuest~\cite{ranasinghe-etal-2020-transquest}, in both Independent Language-Pair Training and Unified Multilingual Training settings.  For comparison, we use the COMET model~\cite{rei-etal-2023-scaling}, which is fine-tuned on low-resource language pairs (mentioned in the section~\ref{subsec:datasets}) utilizing the pre-trained encoder transformer XLM-R-XL~\cite{DBLP:journals/corr/abs-2105-00572}.  We chose to restrict the investigation to zero-shot, in-context learning and adapter fine-tuning. Approaches which use continual pre-training are not within the scope of this investigation due to their computational cost, leaving them for future work.
\paragraph{Zero-shot and ICL scenarios}  We utilize the vLLM framework~\cite{kwon2023efficient} to perform our experiments. For all our zero-shot and ICL experiments, we experimented with the default temperature value of 0.85 and also the value of 0. The temperature value of 0 provided a more stable and consistent output. The input sequence length was set to 1024 for zero-shot inference and 4096 for ICL inference.

\paragraph{Instruction fine-tuning} We used the LLaMA-Factory framework~\cite{zheng2024llamafactory} for fine-tuning experiments, leveraging its prompt formatting capabilities. For efficient tuning, we applied LoRA ~\cite{hu2021lora}, focusing on the query and value projection layers of transformers, which proved the most effective in reducing computational cost and memory usage. This approach consistently provided reliable outputs, making these layers our choice for fine-tuning throughout the experiments.
% We set the LoRA rank to 64 to balance model performance with computational efficiency. A higher rank enhances adaptation but demands more memory and computing power. Using 4-bit quantization, we reduced model size and accelerated inference, decreasing memory usage and speeding up computations, although with a slight impact on accuracy~\cite{dettmers2023qlora}. Additionally, we used 16-bit floating-point precision (fp16) to further reduce memory requirements and increase training speed~\cite{micikevicius2018mixed}, allowing us to handle larger models and batch sizes within the same memory constraints. 
We set the LoRA rank to 64, as higher ranks improve adaptation but increase resource demands. To reduce memory usage and speed up inference, we applied 4-bit quantization, with a slight trade-off in accuracy~\cite{dettmers2023qlora}, and used 16-bit floating-point precision (fp16) to enable larger models and batch sizes within the same memory limits~\cite{micikevicius2018mixed}. 

We conducted fine-tuning experiments in two settings:  \textbf{Unified Multilingual Training (UMT)}, we combined training data from 8 low-resource language pairs (En→{Gu, Hi, Mr, Ta, Te} and {Et, Ne, Si}→En) and performed inference using language-specific test sets; \textbf{Independent Language-Pair Training (ILT)}, we fine-tuned separate models for each language pair, using individual training data and performing inference with corresponding test sets to evaluate the results. All the AG prompt data\footnote{\href{https://huggingface.co/datasets/ArchSid/QE-DA-datasets/tree/main}{huggingface.co/datasets/ArchSid/QE-DA-datasets/}} used for Instruction Fine-Tuning and evaluation, along with the fine-tuned models, have been publicly released on the HuggingFace platform (Appendix ~\ref{app:trained_models}).
% For reproducibility, we release prompt-variant datasets, code and models accrued during our investigation publicly.

\subsection{Evaluation \& Metrics}

We primarily use Spearman's correlation ~\cite{sedgwick2014spearman} between the DA mean (averaged human-annotated DA scores from three annotators) and predictions as our evaluation metric. Additionally, Pearson's correlation ~\cite{cohen2009pearson} and Kendall's Tau correlation ~\cite{lapata-2006-automatic} are calculated (see Appendices: ~\ref{app:zeroshot_all}, ~\ref{app:icl_all}, ~\ref{app:mono_all}, ~\ref{app:multi_all}).

% We primarily use Spearman's correlation~\cite{sedgwick2014spearman} for evaluation. Additionally, Pearson's correlation~\cite{cohen2009pearson} and Kendall's Tau~\cite{lapata-2006-automatic} scores are also generated as shown in the Appendices:~\ref{app:zeroshot_all},~\ref{app:icl_all},~\ref{app:mono_all}~\ref{app:multi_all}.

% The predicted outputs from our models contained additional text alongside the predicted score. We use \textit{regular expressions} to extract the predicted DA score from the output. In the zero-shot and ICL experiments using LLMs, some outputs did not include a score in the predicted text, and these instances were excluded from the correlation computation. Details regarding the number of rows excluded from the analysis are provided in Appendix ~\ref{app:zeroshot_all} \& ~\ref{app:icl_all}. However, this problem is mitigated after instruction fine-tuning, and all inferenced instances from the fine-tuned model successfully predicted a score.

The predicted outputs from our models contained extra text alongside the predicted DA score, which we extracted using regular expressions. In the zero-shot and ICL experiments, some outputs lacked a score, and those cases were excluded from the correlation analysis (see Appendices ~\ref{app:zeroshot_all} \&~\ref{app:icl_all}). However, this problem is mitigated after instruction fine-tuning where all inferenced instances predicted a score.

% \paragraph{Statistical Significance} We conducted a two-tailed paired T-test to determine statistical significance between the human-annotated and predicted DA scores. The p-value obtained from this test indicates the statistical significance with $p$ threshold $< 0.05$. For brevity, we only indicate statistically insignificant outcomes with $\dagger$, in Tables~\ref{tab:zero_few} and~\ref{tab:multilingual_result}. All other results are below the threshold with many $< 0.01$  and $< 0.001$ thresholds, indicating a high statistical significance.

\paragraph{Statistical Significance} We performed a two-tailed paired T-test to assess statistical significance between human-annotated and predicted DA scores, using a significance threshold of $p$ $< 0.05$. Statistically insignificant results are marked with $\dagger$ in Tables~\ref{tab:zero_few} and~\ref{tab:multilingual_result}; most other results showed high significance, with $p < 0.01$ or $p < 0.001$.

\section{Results }\label{sec:results}

\begin{table*}[ht]
\centering
% \small
\tiny
\resizebox{\textwidth}{!}{
\begin{tabular}{ccccccc}
\hline
\textbf{Lang-pair} & \textbf{Gemma-7B} & \textbf{Llama-2-7B} & \textbf{Llama-2-13B} & \textbf{OC-3.5-7B} & \textbf{TransQuest} & \textbf{CometKiwi} \\ 
\hline
\multicolumn{7}{c}{\textbf{Unified Multilingual Training (UMT) Setting}} \\
\hline
En-Gu & \underline{0.566} & 0.461 & 0.465 & 0.554 & 0.630 & \textbf{0.637} \\
En-Hi & 0.449 & 0.332 & 0.322 & \underline{0.458} & 0.478 &\textbf{0.615} \\
En-Mr & \underline{0.551}\textsuperscript{$\dagger$} & 0.516\textsuperscript{$\dagger$} & 0.505 & 0.545\textsuperscript{$\dagger$} & \textbf{0.606} & 0.546 \\
En-Ta & 0.502 & 0.464 & 0.471 & \underline{0.509} & 0.603 & \textbf{0.635} \\
En-Te & 0.242 & 0.258 & 0.258 & \underline{0.267} &  \textbf{0.358} & 0.338 \\ \hdashline
Et-En & \underline{0.728} & 0.636 & 0.655 & 0.678 & 0.760 & \textbf{0.860} \\
Ne-En & \underline{0.650} & 0.519 & 0.565 & 0.607 & 0.718 & \textbf{0.789} \\
Si-En & 0.455 & 0.395 & 0.403\textsuperscript{$\dagger$} & \underline{0.481}\textsuperscript{$\dagger$} & 0.579 & \textbf{0.703} \\
\hline
\multicolumn{7}{c}{\textbf{Independent Language-Pair Training (ILT) Setting}} \\
\hline
En-Gu & 0.440 & 0.214 & 0.421 &  \underline{0.520} & \textbf{0.653}& -  \\
En-Hi & 0.375 & 0.282 & 0.336 & \underline{\textbf{0.474}} & 0.119 & -  \\
En-Mr &  \underline{0.557} & 0.509\textsuperscript{$\dagger$} & 0.501 & 0.554\textsuperscript{$\dagger$} & \textbf{0.629}& -  \\
En-Ta & 0.475 & 0.375 & 0.441 & \underline{\textbf{0.509}} & 0.303& -  \\
En-Te & 0.217 & 0.263 & 0.261 &\underline{\textbf{0.271}} & 0.087& -  \\ \hdashline
Et-En & 0.648 & 0.589 & 0.598 &  \underline{0.652} & \textbf{0.806}& -  \\
Ne-En & 0.612 & 0.497 & 0.543\textsuperscript{$\dagger$} & \underline{0.614} & \textbf{0.746}& -  \\
Si-En & 0.387 & 0.332 & 0.346 &  \underline{0.441} & \textbf{0.581}& -  \\
\hline
\end{tabular}
}
\caption{\small{Spearman correlation ($\rho$) scores between the predicted and mean DA scores for \textit{UMT} and \textit{ILT} fine-tuning. For both settings exclusively, scores underlined represent best amongst LLMs, and scores in boldface indicate overall best scores amongst both LLMs and encoder-based models. (\textsuperscript{$\dagger$}) denotes the statistically insignificant results ($p > 0.05$). The dashed line separates language pairs with English as the target.}}  
\label{tab:multilingual_result}
\end{table*}

Table~\ref{tab:zero_few} presents results from the zero-shot and ICL scenarios. Our proposed AG prompt achieved the highest scores in the zero-shot setting for most language pairs, with the exception of En to\{Gu, Hi, Te\}. For En to \{Hi, Te\} the AG prompt scores were very close to those of the best scores, indicating the AG prompt's strength across the majority of language pairs.  Notably, the OpenChat model attained the highest correlation scores for all language pairs in the zero-shot experiment. 

% Given the strong performance of the AG prompt in zero-shot experiments, our investigation in the ICL setting focused exclusively on the AG prompt. In the ICL setting, 4 language pairs (En-Gu, En-Mr, En-Te, Ne-En) demonstrated optimal performance with the 7-ICL configuration. Three language pairs (En-Hi, En-Et, Si-En) achieved the best results with the 5-ICL setting, while only one language pair (En-Ta) showed the highest performance with the 3-ICL setting. The OpenChat model consistently achieved the highest correlation scores across all low-resource language pairs in every ICL configuration.  It is also noteworthy that the zero-shot and ICL performance for the Et-En, Ne-En, and Si-En language pairs are comparatively higher than those for the other Indic language pairs.

Given the AG prompt's strong zero-shot performance, our ICL investigations focused solely on it. In the ICL setting, 4 language pairs (En-Gu, En-Mr, En-Te, Ne-En) performed best with 7-ICL, 3 language pairs (En-Hi, En-Et, Si-En) with 5-ICL, and 1 language pair (En-Ta) with 3-ICL. OpenChat consistently achieved the highest correlation scores across all low-resource pairs, with Et-En, Ne-En, and Si-En outperforming other English-Indic pairs in both zero-shot and ICL.

In Appendix Tables~\ref{tab:zeroshot_gemba},~\ref{tab:zeroshot_iit}, and~\ref{tab:zeroshot_sai}, for the zero-shot setting, we note that the number of dropped rows for the TE prompt is the highest whereas the same when using AG prompts is the lowest, likely because AG prompt specifies the score ranges explicitly.
 
\paragraph{UMT Setting} As shown in Table~\ref{tab:multilingual_result}, the OpenChat model achieved the highest correlation scores for En to \{Hi, Ta, Te, Si\} while Gemma obtained the highest correlation scores for En-\{Gu,Mr\} and \{Et, Ne\}-En. However, compared to instruction fine-tuned LLMs, the fine-tuned encoder-based models (TransQuest, CometKiwi) consistently achieved significantly higher correlations among all low-resource language pairs.
\paragraph{ILT Setting} As shown in Table~\ref{tab:multilingual_result}, OpenChat obtained the best Spearman scores among other LLMs for all the language pairs except En-Mr. Unlike UMT fine-tuning where only pre-trained encoders gave the best result, ILT fine-tuned LLMs achieve the highest results for En to \{Hi, Ta, Te\} in this setting, where Tamil and Telugu languages are from the Dravidian family which are considered extremely low-resource in terms of pre-training data distribution for LLMs.  

Comparing ILT and UMT setting results, the UMT performs better for most low-resource language pairs. This suggests that incorporating diverse linguistic data enhances the model's ability to generalize and accurately evaluate translations across various low-resource languages. Considering the overall best results, fine-tuned encoder-based models demonstrate the best performance.

\section{Discussion}\label{sec:discussion}

\paragraph{Zero-shot- }In comparison to the GEMBA and TE prompts, the AG prompt demonstrated the best overall performance in zero-shot experiments with LLMs across the majority of language pairs. This indicates that in the absence of training data, the additional context provided in the AG prompt—acting as an annotation guide, enhances the effectiveness of LLM-based quality estimation more effectively than LLMs functioning as translation evaluators (TE template) or simply assigning scores based on a straightforward request like in the GEMBA prompt. The structured guidelines in the AG prompt offer a clearer framework for evaluating translation quality, which supports more accurate scoring in zero-shot settings.

% \paragraph{ICL - }Comparing the results from zero-shot and ICL experiments, the ICL setting yielded the highest correlation scores for the majority of language pairs (En-Gu, En-Te, Et-En, Ne-En, Si-En). This indicates adding examples to the prompts could enhance the knowledge of the LLMs to predict the quality of the translation. Moreover, the overall ICL experiment results suggest that the effect of increasing the number of examples in ICL is not uniform across all language pairs and models (See Appendix~\ref{app:all_setting_spearman}). When the number of examples in the ICL prompts was increased, the En-Gu and Ne-En language pairs with the Gemma-7B model, as well as the En-Mr and Ne-En language pairs with the OpenChat model, consistently showed improved performance. However, for other language pairs and models, the performance gains were not always evident, suggesting that increasing the number of examples does not necessarily lead to better results.

\paragraph{ICL-} Outperformed zero-shot for most language pairs (En-Gu, En-Te, Et-En, Ne-En, Si-En), suggesting that adding examples improves LLMs' ability to predict translation quality. However, the effect of increasing examples varied across language pairs and models (see Appendix~\ref{app:all_setting_spearman}). When the number of examples in the ICL prompts was increased, the En-Gu and Ne-En language pairs with the Gemma-7B model, as well as the En-Mr and Ne-En language pairs with the OpenChat model, consistently showed improved performance. However, for other language pairs and models, the performance gains were not always evident, suggesting that increasing the number of examples does not necessarily lead to better results.

% However, we observed a significant improvement in the correlation scores with LLMs when progressing from zero-shot to fine-tuning (See Appendix~\ref{app:all_setting_spearman}) when compared to zero-shot to ICL. This suggests that instruction fine-tuning LLMs with task-specific data enhances their performance more effectively than providing detailed examples within the prompts.

\paragraph{Fine-tune -}We observed a notable improvement in correlation scores when moving from zero-shot to fine-tuning, compared to zero-shot to ICL (Appendix~\ref{app:all_setting_spearman}). This indicates that instruction fine-tuning with task-specific data is more effective than providing detailed examples in prompts.
% In fine-tuning experiments, UMT pre-trained encoder-based models achieved the highest results surpassing the LLMs. Comparing the size (number of parameters) and the disk space utilisation of the models, LLMs are significantly larger and contain more parameters than pre-trained encoder-based language models (See Appendix~\ref{app:model_size}). 
In fine-tuning experiments, pre-trained encoder-based models with UMT settings outperformed LLMs. Despite this, LLMs are significantly larger in size and contain more parameters compared to pre-trained encoder models (Appendix~\ref{app:model_size}).
% However, LLMs can perform various NLP tasks, and if already deployed, can prove to be useful for translation evaluation, given decent performance for some low-resource language pairs.  Additionally, LLMs are not specifically trained for regression tasks, whereas pre-trained encoders are fine-tuned to predict a score. This fundamental difference may contribute to the lower performance of LLMs as QE models. The OpenChat model consistently outperformed other LLMs across most experimental settings and language pairs, given sufficient context as annotation guidelines for low-resource languages.
While LLMs can handle various NLP tasks and show decent performance in translation evaluation for some low-resource language pairs, they are not specifically trained for regression tasks like pre-trained encoders. This difference likely contributes to LLMs' lower performance in QE. Notably, the OpenChat model consistently outperformed other LLMs when provided with sufficient context as annotation guidelines.

\begin{figure*}
\centering
\begin{minipage}{.44\textwidth}
  \centering
  \includegraphics[width=\linewidth]{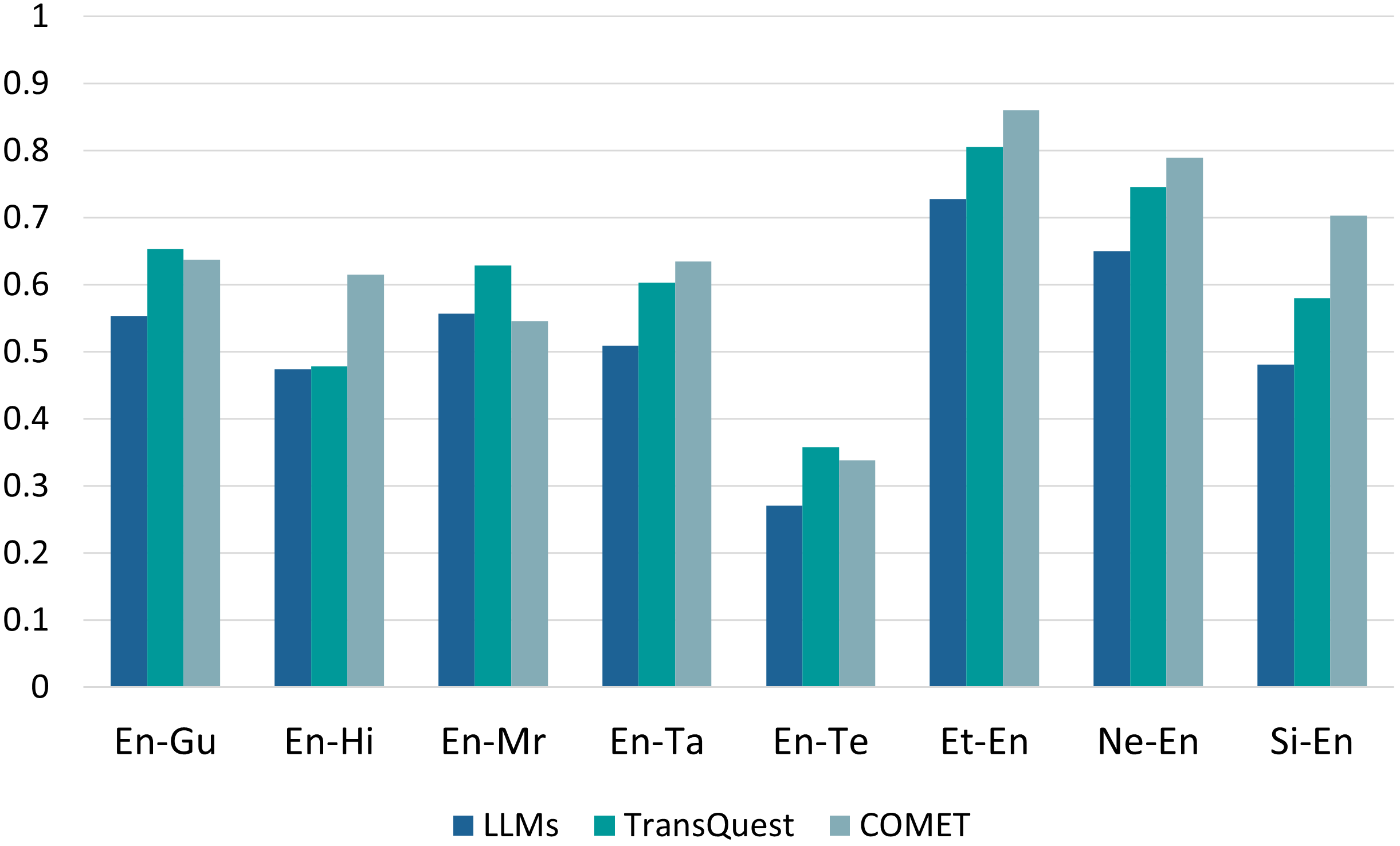}
  \caption{\small{Best fine-tuned performance (Spearman) for LLMs \textit{vs.} TransQuest-InfoXLM \textit{vs.} COMET}}
  \label{fig:bestscores}
\end{minipage}
\quad
\begin{minipage}{.485\textwidth}
  \centering
  \includegraphics[width=\linewidth,trim={0.9cm 1cm 1.5cm 2cm},clip]{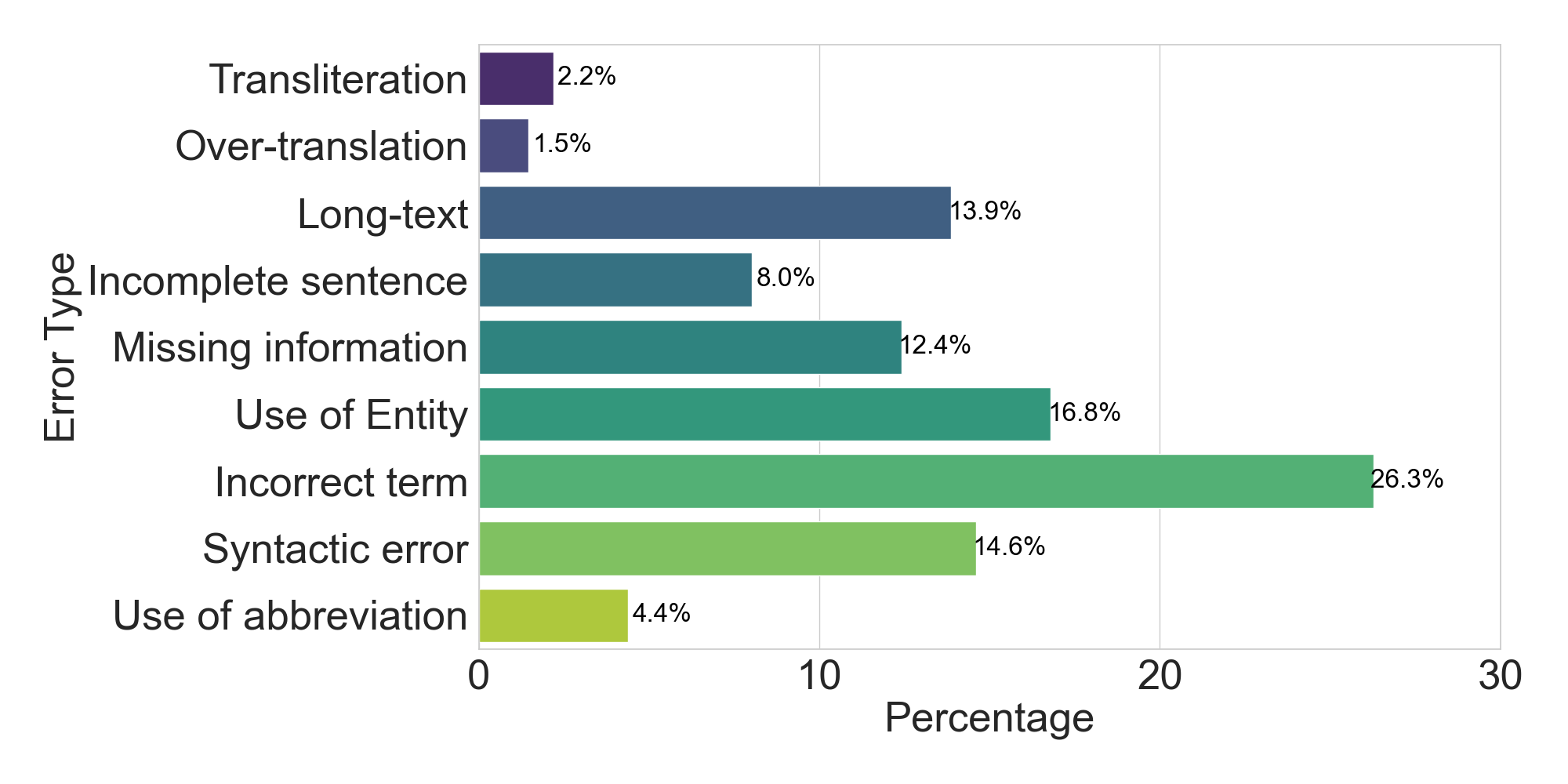}
  \caption{\small{Error types and their percent contribution.}}
  \label{fig:error_analysis_bargraph2}
\end{minipage}%

\end{figure*}

A noteworthy observation is that English, when used as the target language in machine translation, consistently achieved higher correlation scores for QE in zero-shot and ICL experiments with LLMs. Similarly, Figure~\ref{fig:bestscores}, which highlights \textit{setting-agnostic} best performance for fine-tuned LLMs \textit{vs.} TransQuest-InfoXLM \textit{vs.} COMET, shows enhanced performance with English as the target language and the data distribution for other language pairs is a concern for most pre-training setups~\cite{uthus-etal-2023-mlongt5}, including those of encoder models. This observation is in line with the study of ~\citet{nguyen-etal-2024-democratizing} and indicates that language models are likely more proficient when English is the target language, which consequently leads to enhanced performance by LLMs and encoders, and poses a question on multilingual claims made by LLM releases. 

Figure~\ref{fig:bestscores} indicates LLMs are outperformed for most language pairs by TransQuest-based and COMET models. Interestingly, for the only language pair where LLMs match COMET performance, En-Mr, the results are statistically insignificant. Among the Indic-target language pairs, En-Mr shows a consistently higher correlation, but statistically insignificant in most cases (Table~\ref{tab:multilingual_result}) across both settings. In the UMT setting, this could be an outcome of imbalanced data distribution since En-Mr has a significantly large training set, but we have similarly insignificant outcomes from the ILT setting as well. 
Our work indicates that LLM-based adapters may not perform as well as encoder-based models. Investigating larger variants may produce better performance but smaller segment-level encoder-based QE models render this direction inefficient. Further, due to the black-box nature of Transformer-based language models, we resort to a tokenization analysis which reveals likely explanations for their QE performance.

\paragraph{Tokenization analysis}\label{sec:tokenization} To explore the reasons behind the better performance of fine-tuned pre-trained encoders over LLMs in reference-less QE tasks, we conducted an analysis of token counts generated by LLMs and pre-trained encoders, such as TransQuest's InfoXLM and CometKiwi's XLM-R-XL. For comparison, high-resource language pairs from the WMT22 test data~\cite{zerva-etal-2022-findings} were included to assess tokenization differences across languages with varying resources.

\begin{figure*}
\centering
\includegraphics[width=0.99\textwidth, keepaspectratio]{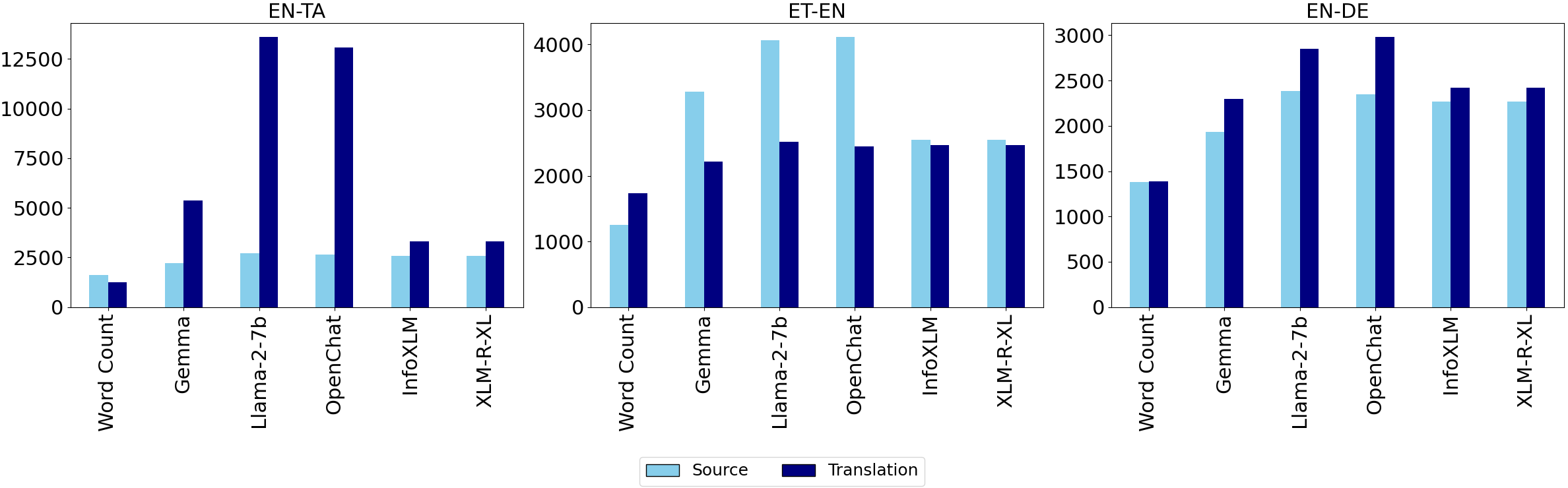}
\caption{
\small{ The graphs compare the original word counts with the model-generated token counts for selected inputs, as described in Section~\ref{sec:tokenization}. This comparison includes both low-resource language pairs (En-Ta, Et-En) and a high-resource language pair (En-De). A detailed image covering all language pairs is provided in the Appendix~\ref{app:token}.}}
\label{fig:3-lang_token} % Give a unique label
\end{figure*}

% We perform analysis to identify why fine-tuned pre-trained encoders outperform LLMs in reference-less quality estimation tasks by examining the token counts generated by LLMs and pre-trained encoders (TransQuest's InfoXLM and CometKiwi's XLM-R-XL). For comparison purposes, we incorporated high-resource language pairs from the WMT22 test data~\cite{zerva-etal-2022-findings} for this analysis.

% We selected 100 sentences for each language pair from our test set. Then we created a tokenization pipeline for each language model. We input the source text and the translation from each language pair as the input to this pipeline and observe the token count from each model. Figure ~\ref{fig:3-lang_token} shows the word count of source and target translation sentences and the token counts of these sentences with each model for three language pairs. We observe a slight difference in token counts between Llama-2-7B and OpenChat-3.5, despite both models using the same tokenizer. The tokenization analysis for all the language pairs is present in the Appendix~\ref{app:token}.  

We selected 100 sentences per language pair from our test set and created a tokenization pipeline for each model. Both source and translation texts were input to observe token counts. Figure ~\ref{fig:3-lang_token} shows word and token counts for three language pairs, revealing slight differences between Llama-2-7B and OpenChat-3.5 despite using the same tokenizer. The tokenization outcomes for all language pairs are detailed in Appendix~\ref{app:token}. The token counts generated by LLMs (Gemma, OpenChat, Llama) for low-resource non-English languages significantly deviate from the original word counts, while pre-trained encoders like InfoXLM and XLMR-XL show smaller discrepancies. Rich morphological languages like Marathi, Tamil, and Telugu, which feature agglutinative\footnote{A grammatical process in which words are composed of a sequence of morphemes (meaningful word elements), each of which represents not more than a single grammatical category} phrases, and Hindi, which includes compounding, experience skewed tokenization, affecting semantic matching between source and translation (Appendix~\ref{app:token}). In contrast, for English, the tokenized count closely matches the word count, regardless of the model used. This highlights the need for improved tokenization strategies for cross-lingual semantic matching with LLMs for low-resource languages to enhance performance on the QE task.

We also identified that the Et-En language pair consistently achieved the highest performance across all experimental settings. As illustrated in the Appendix~\ref{app:token}, the difference between the token counts generated by language models \textit{vs.} the original word counts is evidently smaller than that observed for other low-resource languages. This holds true even for LLMs as well. This reduced tokenization discrepancy, likely due to both languages (En and Et) using the Latin alphabet, may explain why Et-En performs better in all experimental settings. 
% Additionally, it is evident that the Llama models exhibit the highest difference in token counts for low-resource non-English languages compared to the original word counts. Correspondingly, Llama2 variants showed lower performance in all the experiments compared to other LLMs. 

Looking at the tokenization for high-resource non-English languages (see Appendix~\ref{app:token}), it can be seen that the language pairs En-De (German) and Ro (Romanian)-En exhibit limited disparity in the number of tokens from the original word count, and for En-Zh (Chinese) it significantly higher. De/Ro uses Latin-based scripts too. This analysis suggests that English and other Latin-script-based languages, benefit from more efficient tokenization in language models, which leads to improved performance in tasks like QE. In contrast, other languages, exhibit greater disparities in token counts, indicating the need for more advanced tokenization strategies with LLMs to enhance performance. This underscores the importance of developing better tokenization methods to ensure equitable model performance across different language pairs.
\paragraph{Error Analysis} \label{sec:erroranalysis} We conducted an error analysis using the top-performing model, OpenChat, focusing solely on the En-Ta language pair due to native speaker availability. The purpose of this analysis was to identify the underlying reasons for significant deviations in predicted DA scores from the ground truths, aiming to understand what factors in the input contribute to inaccurate predictions. From the model's predictions, we selected 100 sentences with the highest deviations between predicted and human-annotated DA scores. Figure~\ref{fig:error_analysis_bargraph2} presents the identified error types and their occurrence percentages. The annotated error types are based on the Multidimensional Quality Metric Error typology~\citep{lommel-etal-2014-using}.

A significant portion of errors, such as \textit{Incorrect term} (26.3\%), \textit{Use of Entity} (16.8\%), and \textit{Syntactic error} (14.6\%), suggests that the model struggles with accurately understanding the contextual appropriateness in the translations to predict the DA score. This can be attributed to the inherent challenges in capturing the nuances and complexities of language, especially for low-resource languages where the training data may be insufficient or lacks diversity. The \textit{Long-text} (13.9\%) and \textit{Incomplete sentence} (8\%) errors indicate difficulties in maintaining coherence and completeness in translation, which are crucial for accurate QE.  \textit{Missing information} (12.4\%) which highlights the challenge of ensuring the completeness of the sentence and \textit{Transliteration} errors (2.2\%) highlighting the challenges of understanding the conversion of phonetic elements also seem to be important for accurate quality estimation. Finally \textit{Use of abbreviation} errors (4.4\%) suggest that the model is unlikely to have seen domain-specific terminology, which requires domain-specific training data for better quality estimation.

\section{Conclusion and Future Work}\label{sec:conclusion}

% This paper conducts an investigation into large language models based on reference-less quality estimation for low-resource language pairs. We reproduce results using existing prompting strategies and propose two modifications where our proposed AG prompt which provides additional context based on summarized human annotation guidelines performs the best in the zero-shot scenario. Integrating examples into the ICL configuration with our proposed AG prompt led to an improvement in LLM performance for QE for most of the language pairs compared to the zero-shot scenario. Instruction fine-tuning with parameter-efficient techniques significantly enhances the performance of LLMs by reducing the required memory and delivering comparable results for some language pairs. However, they are outperformed by pre-trained encoder-based supervised regression models, in line with the findings of~\citet{vandan-etal-2023-towards}.

This paper investigates reference-less quality estimation for low-resource language pairs using large language models. We reproduce results with existing SOTA prompts and propose a new AG prompt, which performs best in zero-shot settings. Further experiments with ICL and instruction fine-tuning settings are performed with AG prompt which achieves closer performance with the pre-trained encoder-based approaches.

Our findings indicate how LLM-based QE can be challenging for morphologically richer languages without much data in the pre-training stage. Based on our findings, we highly recommend the addition of QE datasets to LLM evaluation task suits given the significant cross-lingual challenge posed by this task. We perform a detailed tokenization analysis which highlights that cross-lingual machine understanding for low-resource languages needs to be addressed at the stage of tokenization~\cite{remy2024trans}, and within pre-training data~\cite{petrov2024language}. Additionally, error analysis highlights significant challenges in handling context, syntax, and domain-specific terms, suggesting that further refinement in model training and adaptation is necessary. In the future, we aim to employ regression head-based adapters within the LLM pipeline for QE, eliminating the challenges in the reliability of extracting the scores from the outputs.

% We also note the challenge of extracting a predicted score from generated text, which poses a question on the reliability of such approaches, and suggest alternate future investigations below.

% In future, we aim to focus on enhancing the contextual understanding of LLMs through improved fine-tuning techniques and including more diverse and extensive training datasets. We also aim to use regression head-based adapters for fine-tuning an LLM pipeline for quality estimation. However, given the pre-training objective of LLMs, a more detailed investigation for a unified evaluation and correction is warranted, likely explaining the errors and reasoning behind the correction.  

\section{Limitations}
% The unavailability of diverse datasets from different domains for low-resource languages limits the generalizability of the results. While our study demonstrates the potential of using LLMs with parameter-efficient fine-tuning techniques for quality estimation, the models still demand considerable processing power and memory, which may not be readily available to all researchers. 

Our results are based on a limited number of LLMs, primarily smaller than 14 billion parameters, due to the constraints imposed by our computational resources. All experiments were conducted using only one GPU (NVIDIA A40 40G), which required significant time for instruction fine-tuning and inference across several language pairs. Additionally, our study was limited to open-source LLMs. 

The availability of human-annotated DA scores for low-resource languages is limited to the eight language pairs included in this study and our analysis is constrained to these specific datasets. In the future, we aim to expand our study to include datasets where the source and translated languages are reversed, provided such datasets become available.
\bibliography{custom, anthology}
\clearpage % Ensure the appendix starts on a new page
\onecolumn % Switch to one-column format if you want the appendix in one column
\appendix

\section{Appendix: In-context learning prompt} \label{app:icl}

 \begin{figure*}[ht]
 \centering
 \includegraphics[width=0.8\textwidth, keepaspectratio]{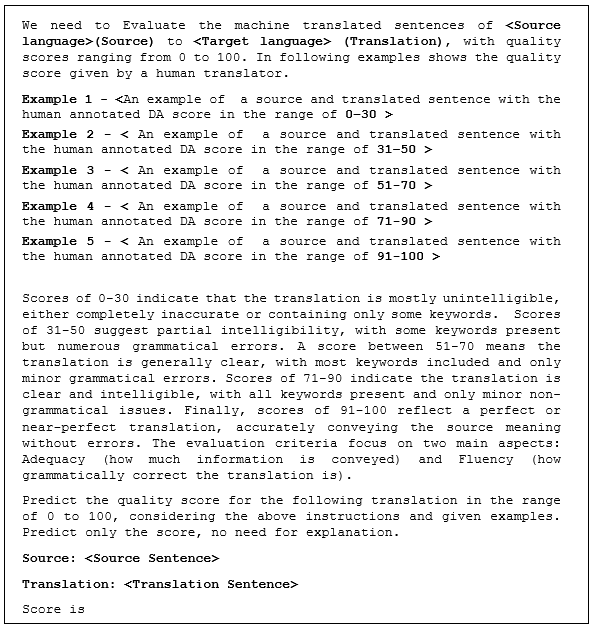}
\caption{ Our proposed AG prompt for in-context learning.}
\label{fig:fewshot} % Give a unique label
\end{figure*}
\clearpage % Ensure the appendix starts on a new page

\section{Appendix: Other prompts} \label{app:other_prompts}

 \begin{figure*}[ht]
 \centering
 \includegraphics[width=0.9\textwidth, keepaspectratio]{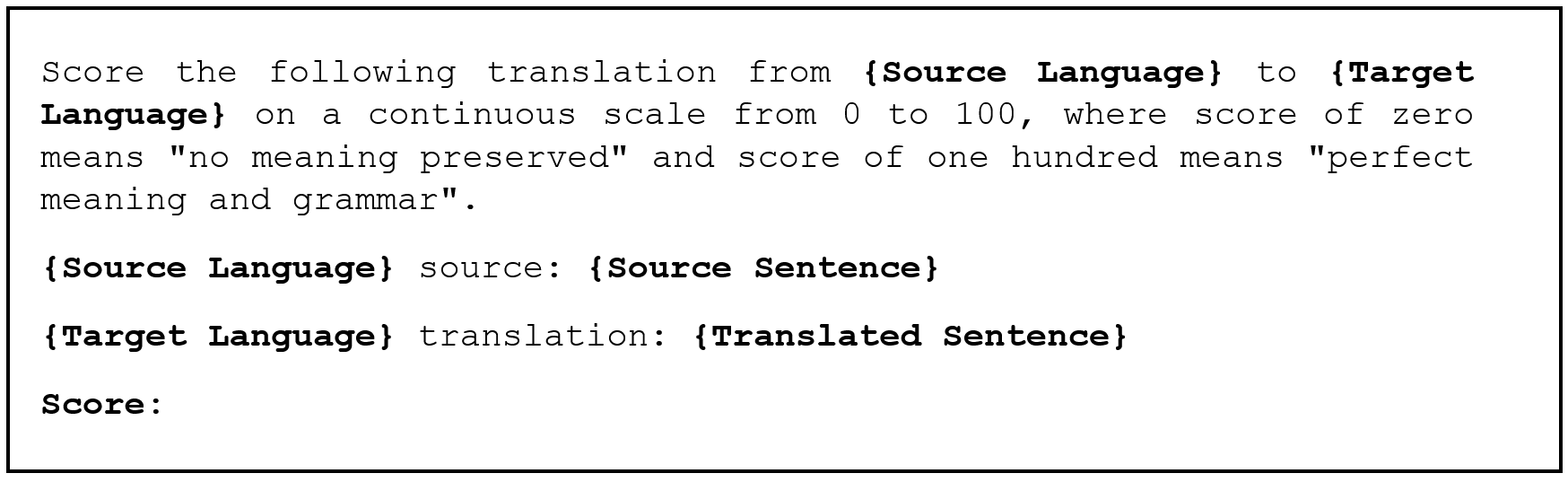}
\caption{ GEMBA prompt~\cite{kocmi-federmann-2023-large}}
\label{fig:gemba} % Give a unique label
\end{figure*}

The \textbf{GEMBA } prompt is part of the GEMBA (GPT Estimation Metric Based Assessment) method, which uses GPT-based language models to evaluate translation quality. The GEMBA  prompt evaluates translation quality by scoring each translation segment on a continuous scale from 0 to 100.

 \begin{figure*}[ht]
 \centering
 \includegraphics[width=0.88\textwidth, keepaspectratio]{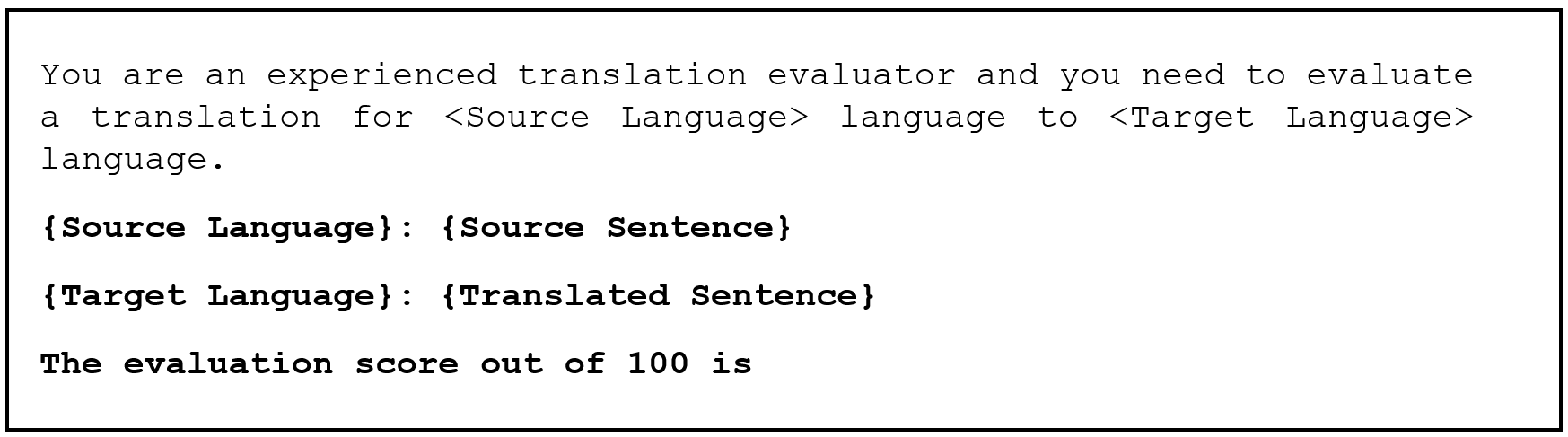}
\caption{ TE prompt~\cite{vandan-etal-2023-towards}}
\label{fig:te-prompt} % Give a unique label
\end{figure*}

The \textbf{TE (Translation Evaluator)} prompt instructs the model to act as an experienced translation evaluator, explicitly presenting the source language, source text, target language, and translated text. The prompt concludes with the model assigning a score out of 100 to the translation, indicating its quality.

 % \clearpage % Ensure the appendix starts on a new page

\section{Appendix: Train and test data splits} \label{app:DA_dataset}

\begin{table}[ht]

\centering
\small % Reduces the font size
\setlength{\tabcolsep}{15pt} % Reduces the space between columns
\begin{tabular}{@{}lcc@{}}
\hline
% \multicolumn{4}{c} \\
% \multicolumn{4}{c} {DA Score} \\
\toprule

Lang. & Train  & Test \\ 
\hline
\\

English - Gujarati (En-Gu) & 7000 & 1000 \\ \\
English - Hindi (En-Hi) & 7000  & 1000\\ \\
English - Marathi (En-Mr) & 26 000  & 699 \\ \\
English - Tamil (En-Ta) & 7000  & 1000 \\ \\ 
English - Telugu (En-Te) & 7000  & 1000 \\ \\
Estonian - English (Ne-En) & 7000 & 1000\\ \\
Nepalis - English (Ne-En) & 7000  & 1000\\ \\
Sinhala - English (Si-En) & 7000  & 1000 \\ \\
\bottomrule
\end{tabular}
\caption{The dataset splits of translation datasets with human-annotated DA scores utilized in our study. We conducted experiments on 8 low-resource language pairs to evaluate the performance of various models.}

\label{tab:datasets}
\end{table}

\section{Appendix: Train and test data with number of instances in each DA score ranges} \label{app:DA_distribution}

 \begin{figure}[H]
 \centering
 \includegraphics[width=0.9\textwidth, keepaspectratio]{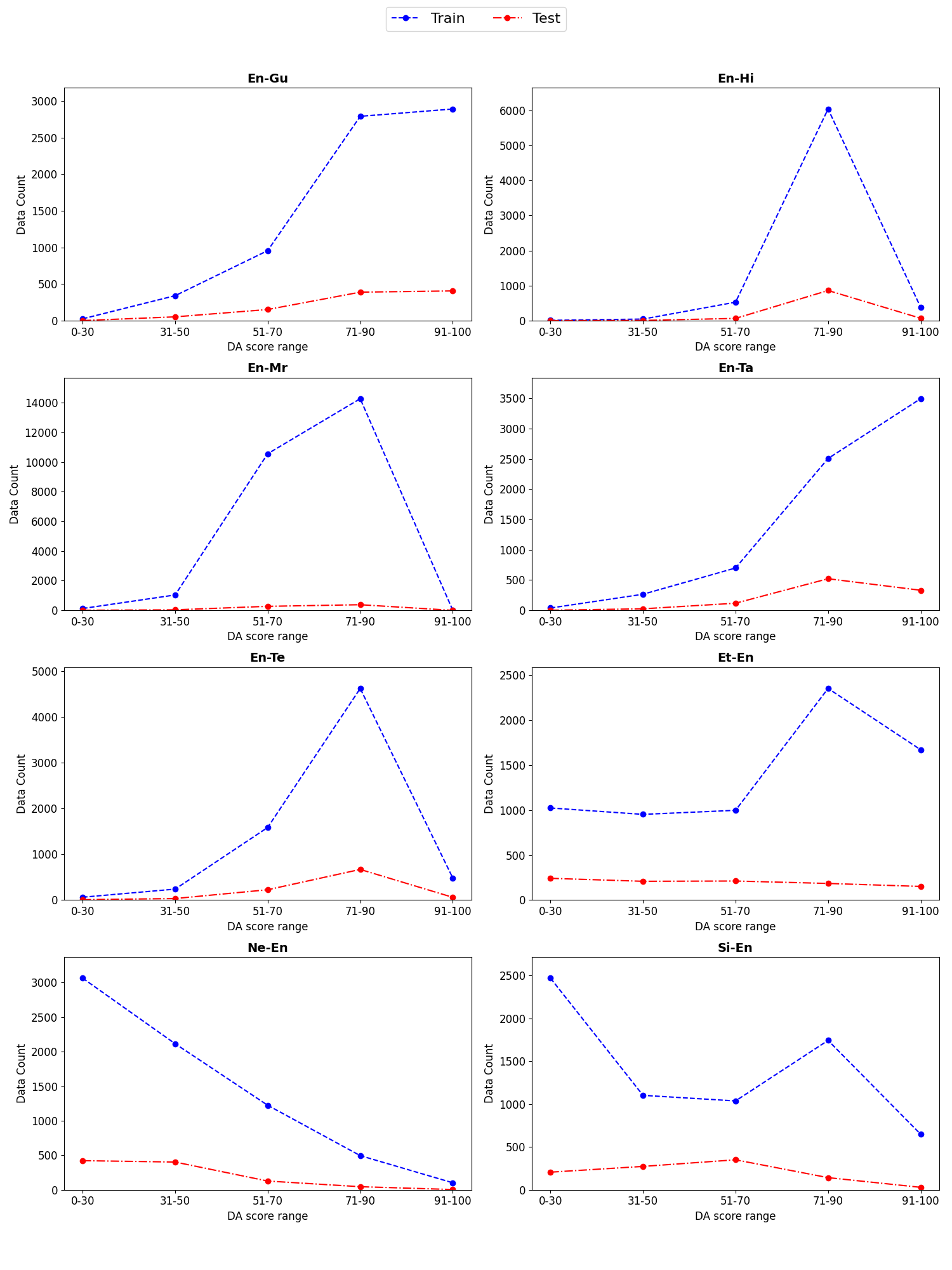}
\caption{ This image shows the number of data belonging to each DA score range of each language pair in the train and test data sets.}
\label{fig:DA_score_ranges} % Give a unique label
\end{figure}

\clearpage

% \section{Appendix: Tokenization with language models} \label{app:token}

%  \begin{figure*}[htbp]
%  \centering
%  \includegraphics[width=0.85\textwidth, keepaspectratio]{latex/fig/ALL_tokenization_word_counts.png}
% \caption{ This figure shows the comparison between original word count for each low resource language pair and the tokenized count from different language models we have used in our experimental study. X-axis shows the tokenization method and Y-axis shows the count of tokens. }
% %\label{fig:token} % Give a unique label
% \end{figure*}
%  \clearpage % Ensure the appendix starts on a new page

\clearpage

\section{Appendix: Tokenization with different language models} \label{app:token}

\begin{figure}[H]
\centering
\includegraphics[width=0.9\textwidth, keepaspectratio]{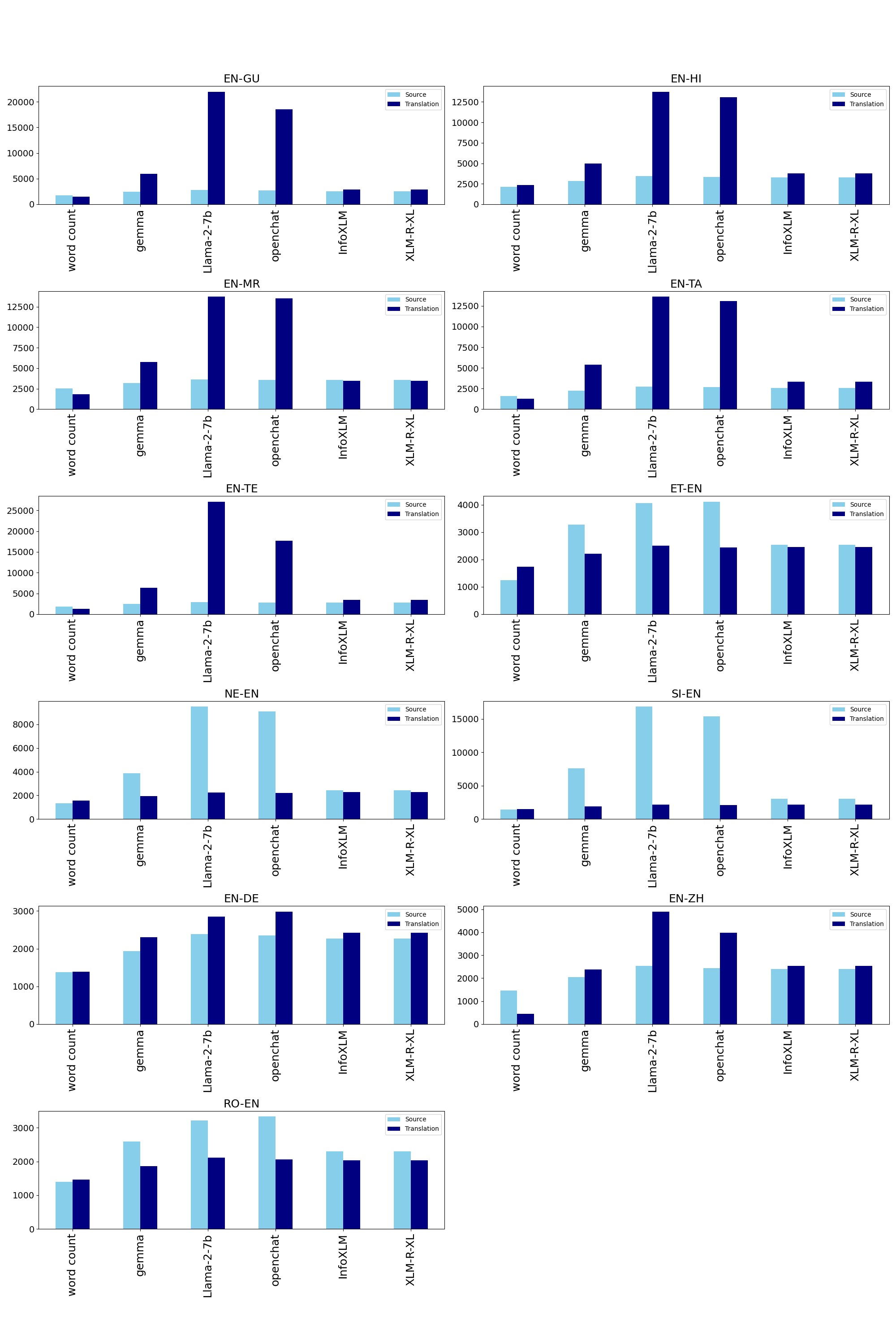}
\caption{ A comparative analysis of the total word count of source and target sentences versus the count of tokens generated by various language models for, both low-resource and high-resource language pairs. The X-axis represents the model name, while the Y-axis indicates the generated token counts. }
%\label{fig:token} % Give a unique label
\end{figure}
\clearpage % Ensure the appendix starts on a new page

\section{Appendix: Zero-shot experiment results with Pearson, Spearman and Kendal's Tau Correlation scores}\label{app:zeroshot_all}

\begin{table}[ht]
    \centering
    \resizebox{\textwidth}{!}{%
    \begin{tabular}{c|cccc|cccc|cccc|cccc}
    \hline
    \multirow{2}{*}{Language pairs} & \multicolumn{4}{c|}{Gemma-7B} & \multicolumn{4}{c|}{Llama-2-7B} & \multicolumn{4}{c|}{Llama-2-13B} & \multicolumn{4}{c}{OC-3.5-7B} \\
    \cline{2-17}
    & r & $\rho$ & $\tau$ & E & r & $\rho$ & $\tau$ & E & r & $\rho$ & $\tau$ & E & r & $\rho$ & $\tau$ & E \\
    \hline

    En-Gu & 0.125 & 0.113 & 0.092 & 0 & 0.015 & 0.006 & 0.005 & 1 & 0.048 & 0.019 & 0.016 & 6 & 0.267 & 0.249 & 0.187 & 1 \\
    En-Hi & 0.154 & 0.131 & 0.106 & 0 & -0.031 & -0.002 & -0.001 & 6 & 0.049 & 0.009 & 0.007 & 6 & 0.315 & 0.254 & 0.188 & 9 \\
    En-Mr & 0.177 & 0.135 & 0.109 & 0 & 0.054 & 0.053 & 0.042 & 0 & 0.103 & 0.115 & 0.088 & 7 & 0.323 & 0.183 & 0.137 & 0 \\
    En-Ta & 0.346 & 0.222 & 0.179 & 0 & 0.034 & 0.067 & 0.054 & 17 & 0.108 & 0.091 & 0.070 & 6 & 0.400 & 0.358 & 0.270 & 4 \\
    En-Te & 0.074 & 0.081 & 0.066 & 0 & 0.005 & -0.016 & -0.013 & 0 & 0.093 & 0.121 & 0.094 & 0 & 0.155 & 0.145 & 0.109 & 0 \\
    Et-En & 0.286 & 0.289 & 0.229 & 1 & 0.173 & 0.168 & 0.129 & 3 & 0.232 & 0.185 & 0.139 & 26 & 0.550 & 0.571 & 0.411 & 3 \\
    Ne-En & 0.261 & 0.261 & 0.199 & 1 & 0.144 & 0.153 & 0.119 & 10 & 0.234 & 0.222 & 0.165 & 11 & 0.476 & 0.448 & 0.320 & 14 \\
    Si-En & 0.272 & 0.193 & 0.150 & 5 & 0.155 & 0.144 & 0.113 & 7 & 0.232 & 0.195 & 0.146 & 5 & 0.439 & 0.417 & 0.299 & 8 \\
   
    \hline
    \end{tabular}
    }
    \caption{The complete results of the zero-shot experiments using large language models and the GEMBA prompt template~\cite{kocmi-federmann-2023-large}. The results include Pearson (r), Spearman ($\rho$), and Kendall's Tau ($\tau$) correlation scores. The column `E' indicates the number of rows excluded because the outputs generated by the large language models did not include a score.}
    \label{tab:zeroshot_gemba}
\end{table}

\begin{table}[ht]
    \centering
    \resizebox{\textwidth}{!}{%
    \begin{tabular}{c|cccc|cccc|cccc|cccc}
    \hline
    \multirow{2}{*}{Language pairs} & \multicolumn{4}{c|}{Gemma-7B} & \multicolumn{4}{c|}{Llama-2-7B} & \multicolumn{4}{c|}{Llama-2-13B} & \multicolumn{4}{c}{OC-3.5-7B} \\
    \cline{2-17}
    & r & $\rho$ & $\tau$ & E & r & $\rho$ & $\tau$ & E & r & $\rho$ & $\tau$ & E & r & $\rho$ & $\tau$ & E \\
    \hline
    En-Gu & -0.094 & -0.102 & -0.085 & 27 & -0.024 & -0.008 & -0.005 & 14 & -0.045 & -0.052 & -0.039 & 102\textsuperscript{*} & 0.180 & 0.117 & 0.085 & 50 \\
    En-Hi & -0.056 & -0.050 & -0.041 & 10 & -0.022 & -0.072 & -0.051 & 28 & 0.047 & 0.056 & 0.041 & 40 & 0.239 & 0.134 & 0.095 & 51 \\
    En-Mr & 0.209 & 0.173 & 0.141 & 12 & 0.070 & 0.070 & 0.048 & 20 & 0.072 & 0.040 & 0.030 & 48 & 0.192 & 0.114 & 0.080 & 34 \\
    En-Ta & -0.017 & -0.037 & -0.030 & 17 & -0.002 & 0.012 & 0.009 & 47 & -0.036 & 0.016 & 0.011 & 143 \textsuperscript{*} & 0.178 & 0.178 & 0.126 & 66 \\
    En-Te & 0.026 & 0.018 & 0.015 & 37 & 0.026 & 0.013 & 0.009 & 26 & -0.007 & 0.010 & 0.008 & 68 & 0.073 & 0.072 & 0.051 & 59 \\
    Et-En & 0.098 & 0.086 & 0.070 & 43 & 0.129 & 0.100 & 0.069 & 2 & 0.157 & 0.146 & 0.107 & 28 & 0.464 & 0.455 & 0.322 & 5 \\
    Ne-En & 0.153 & 0.155 & 0.125 & 90 & 0.142 & 0.100 & 0.070 & 25 & 0.062 & 0.080 & 0.060 & 114\textsuperscript{*} & 0.358 & 0.334 & 0.235 & 94 \\
    Si-En & 0.055 & 0.055 & 0.045 & 20 & 0.134 & 0.129 & 0.091 & 10 & 0.100 & 0.109 & 0.080 & 45 & 0.308 & 0.303 & 0.211 & 43 \\
    \hline
    \end{tabular}
    }
    \caption{The complete results of the zero-shot experiments using large language models and the TE prompt template~\cite{vandan-etal-2023-towards}. The results include Pearson (r), Spearman ($\rho$), and Kendall's Tau ($\tau$) correlation scores. The column `E' indicates the number of rows excluded because the outputs generated by the large language models did not include a score. (*) in the column E indicates that more than 10\% of the total inferences were dropped, which means the results may be considered not trustworthy.  }
    \label{tab:zeroshot_iit}
\end{table}

\begin{table}[ht]
    \centering
    \resizebox{\textwidth}{!}{%
    \begin{tabular}{c|cccc|cccc|cccc|cccc}
    \hline
    \multirow{2}{*}{Language pairs} & \multicolumn{4}{c|}{Gemma-7B} & \multicolumn{4}{c|}{Llama-2-7B} & \multicolumn{4}{c|}{Llama-2-13B} & \multicolumn{4}{c}{OC-3.5-7B} \\
    \cline{2-17}
    & r & $\rho$ & $\tau$ & E & r & $\rho$ & $\tau$ & E & r & $\rho$ & $\tau$ & E & r & $\rho$ & $\tau$ & E \\
    \hline
    En-Gu & -0.034 & -0.079 & -0.059 & 2 & 0.047 & -0.007 & -0.006 & 0 & -0.033 & 0.008 & 0.007 & 0 & 0.159 & 0.164 & 0.132 & 2 \\
    En-Hi & -0.042 & -0.056 & -0.041 & 0 & 0.021 & -0.029 & -0.022 & 0 & 0.051 & 0.069 & 0.055 & 1 & 0.303 & 0.253 & 0.200 & 0 \\
    En-Mr & 0.033 & 0.027 & 0.020 & 3 & 0.097 & 0.059 & 0.046 & 0 & -0.007 & 0.005 & 0.004 & 1 & 0.340 & 0.276 & 0.222 & 0 \\
    En-Ta & 0.026 & -0.002 & 0.000 & 14 & 0.009 & 0.055 & 0.041 & 0 & -0.026 & -0.070 & -0.057 & 1 & 0.367 & 0.363 & 0.290 & 2 \\
    En-Te & 0.072 & 0.065 & 0.048 & 0 & 0.064 & 0.083 & 0.065 & 1 & 0.010 & 0.045 & 0.038 & 0 & 0.129 & 0.121 & 0.095 & 0 \\
    Et-En & 0.077 & 0.098 & 0.071 & 4 & 0.115 & 0.064 & 0.049 & 1 & 0.304 & 0.319 & 0.255 & 1 & 0.615 & 0.619 & 0.470 & 1 \\
    Ne-En & 0.129 & 0.130 & 0.096 & 47 & 0.178 & 0.144 & 0.111 & 1 & 0.283 & 0.303 & 0.236 & 1 & 0.539 & 0.487 & 0.370 & 5 \\
    Si-En & 0.037 & 0.042 & 0.031 & 14 & 0.155 & 0.069 & 0.056 & 5 & 0.267 & 0.238 & 0.185 & 6 & 0.466 & 0.441 & 0.341 & 8 \\
    \hline
    \end{tabular}
    }
    \caption{The complete results of the zero-shot experiments using large language models and the AG prompt template. The results include Pearson (r), Spearman ($\rho$), and Kendall's Tau ($\tau$) correlation scores. The column `E' indicates the number of rows excluded because the outputs generated by the large language models did not include a score.}
    \label{tab:zeroshot_sai}
\end{table}

 \clearpage % Ensure the appendix starts on a new page
\section{Appendix: In-context learning experiment results with Pearson, Spearman and Kendal's Tau correlation scores} \label{app:icl_all}

\begin{table}[ht]
    \centering
    \resizebox{\textwidth}{!}{%
    \begin{tabular}{c|cccc|cccc|cccc|cccc}
    \hline
    \multirow{2}{*}{LP} & \multicolumn{4}{c|}{Gemma-7B} & \multicolumn{4}{c|}{Llama-2-7B} & \multicolumn{4}{c|}{Llama-2-13B} & \multicolumn{4}{c}{OC-3.5-7B} \\
    \cline{2-17}
    & r & $\rho$ & $\tau$ & E & r & $\rho$ & $\tau$ & E & r & $\rho$ & $\tau$ & E & r & $\rho$ & $\tau$ & E \\
    \hline
    En-Gu & 0.010 & -0.005 & 0.003 & 26 & 0.052 & 0.036 & 0.028 & 1 & -0.071 & -0.036 & -0.029 & 1 & 0.202 & 0.223 & 0.174 & 0 \\
    En-Hi & 0.135 & 0.134 & 0.097 & 71 & -0.059 & -0.114 & -0.089 & 1 & 0.009 & -0.023 & -0.019 & 0 & 0.237 & 0.184 & 0.146 & 0 \\
    En-Mr & 0.243 & 0.202 & 0.145 & 111\textsuperscript{*} & 0.130 & 0.120 & 0.089 & 2 & 0.093 & 0.095 & 0.069 & 0 & 0.249 & 0.218 & 0.173 & 0 \\
    En-Ta & 0.106 & 0.122 & 0.089 & 81 & 0.015 & -0.019 & -0.013 & 27 & 0.068 & 0.083 & 0.061 & 1 & 0.252 & 0.337 & 0.270 & 0 \\
    En-Te & 0.104 & 0.092 & 0.068 & 53 & 0.038 & 0.027 & 0.021 & 25 & -0.001 & 0.015 & 0.012 & 0 & 0.083 & 0.152 & 0.124 & 0 \\
    Et-En & 0.233 & 0.226 & 0.162 & 14 & 0.268 & 0.268 & 0.198 & 9 & 0.009 & -0.058 & -0.048 & 4 & 0.590 & 0.613 & 0.459 & 1 \\
    Ne-En & 0.275 & 0.273 & 0.195 & 78 & 0.161 & 0.149 & 0.110 & 5 & 0.322 & 0.340 & 0.266 & 1 & 0.486 & 0.457 & 0.346 & 1 \\
    Si-En & 0.312 & 0.306 & 0.219 & 56 & 0.158 & 0.146 & 0.109 & 19 & 0.150 & 0.018 & 0.013 & 5 & 0.484 & 0.470 & 0.348 & 5 \\
    \hline
    \end{tabular}
    }
    \caption{The complete results of the ICL experiment with 3 examples using our proposed AG prompt template (3-ICL-AG). The results include Pearson (r), Spearman ($\rho$), and Kendall's Tau ($\tau$) correlation scores. `LP'-> Language Pair, `E'-> the number of rows excluded because the outputs generated by the large language models did not include a score. (*) in the column E indicates that more than 10\% of the total inferences were dropped, which means the results may be considered not trustworthy.}
    \label{tab:icl_ag}
\end{table}

\begin{table}[ht]
    \centering
    \resizebox{\textwidth}{!}{%
    \begin{tabular}{c|cccc|cccc|cccc|cccc}
    \hline
    \multirow{2}{*}{LP} & \multicolumn{4}{c|}{Gemma-7B} & \multicolumn{4}{c|}{Llama-2-7B} & \multicolumn{4}{c|}{Llama-2-13B} & \multicolumn{4}{c}{OC-3.5-7B} \\
    \cline{2-17}
    & r & $\rho$ & $\tau$ & E & r & $\rho$ & $\tau$ & E & r & $\rho$ & $\tau$ & E & r & $\rho$ & $\tau$ & E \\
    \hline
    En-Gu & 0.008 & 0.023 & 0.016 & 68 & 0.002 & -0.008 & -0.006 & 1 & 0.087 & 0.095 & 0.070 & 0 & 0.157 & 0.151 & 0.120 & 0 \\
    En-Hi & 0.134 & 0.075 & 0.054 & 32 & 0.002 & -0.022 & -0.016 & 0 & 0.031 & 0.035 & 0.027 & 0 & 0.243 & 0.212 & 0.163 & 0 \\
    En-Mr & 0.218 & 0.164 & 0.119 & 25 & 0.035 & 0.032 & 0.023 & 0 & 0.028 & -0.031 & -0.026 & 0 & 0.256 & 0.226 & 0.181 & 0 \\
    En-Ta & 0.099 & 0.114 & 0.081 & 92 & -0.010 & 0.017 & 0.013 & 1 & 0.095 & 0.193 & 0.146 & 0 & 0.324 & 0.332 & 0.263 & 0 \\
    En-Te & 0.006 & 0.021 & 0.015 & 91 & 0.067 & 0.051 & 0.038 & 0 & 0.023 & 0.073 & 0.057 & 0 & 0.075 & 0.126 & 0.101 & 0 \\
    Et-En & 0.318 & 0.327 & 0.231 & 86 & 0.263 & 0.269 & 0.194 & 2 & 0.461 & 0.438 & 0.322 & 1 & 0.604 & 0.636 & 0.482 & 1 \\
    Ne-En & 0.311 & 0.305 & 0.218 & 98 & 0.203 & 0.189 & 0.138 & 3 & 0.336 & 0.319 & 0.243 & 1 & 0.502 & 0.471 & 0.352 & 1 \\
    Si-En & 0.322 & 0.320 & 0.230 & 37 & 0.123 & 0.243 & 0.186 & 7 & 0.380 & 0.326 & 0.252 & 5 & 0.481 & 0.479 & 0.358 & 5 \\
    \hline
    \end{tabular}
    }
    \caption{The complete results of the ICL experiment with 5 examples using our proposed AG prompt template (5-ICL-AG). The results include Pearson (r), Spearman ($\rho$), and Kendall's Tau ($\tau$) correlation scores. `LP'-> Language Pair, `E'-> the number of rows excluded because the outputs generated by the large language models did not include a score.}
    \label{tab:icl_ag}
\end{table}

\begin{table}[ht]
    \centering
    \resizebox{\textwidth}{!}{%
    \begin{tabular}{c|cccc|cccc|cccc|cccc}
    \hline
    \multirow{2}{*}{LP} & \multicolumn{4}{c|}{Gemma-7B} & \multicolumn{4}{c|}{Llama-2-7B} & \multicolumn{4}{c|}{Llama-2-13B} & \multicolumn{4}{c}{OC-3.5-7B-1210} \\
    \cline{2-17}
    & r & $\rho$ & $\tau$ & E & r & $\rho$ & $\tau$ & E & r & $\rho$ & $\tau$ & E & r & $\rho$ & $\tau$ & E \\
    \hline
    En-Gu & 0.060 & 0.071 & 0.052 & 62 & 0.022 & -0.053 & -0.043 & 3 & -0.093 & -0.108 & -0.082 & 0 & 0.222 & 0.260 & 0.203 & 0 \\
    En-Hi & 0.116 & 0.075 & 0.053 & 64 & -0.088 & -0.176 & -0.139 & 2 & 0.045 & 0.014 & 0.011 & 0 & 0.173 & 0.163 & 0.128 & 0 \\
    En-Mr & 0.256 & 0.167 & 0.126 & 50 & 0.075 & 0.050 & 0.040 & 5 & 0.068 & 0.047 & 0.036 & 0 & 0.277 & 0.251 & 0.201 & 0 \\
    En-Ta & 0.094 & 0.122 & 0.086 & 80 & -0.083 & -0.096 & -0.074 & 0 & -0.059 & -0.004 & -0.004 & 0 & 0.285 & 0.309 & 0.233 & 0 \\
    En-Te & -0.039 & -0.033 & -0.025 & 51 & 0.044 & 0.021 & 0.016 & 0 & -0.009 & -0.028 & -0.023 & 0 & 0.095 & 0.196 & 0.149 & 1 \\
    Et-En & 0.305 & 0.306 & 0.218 & 39 & 0.052 & 0.033 & 0.025 & 1 & 0.198 & 0.169 & 0.125 & 1 & 0.595 & 0.616 & 0.469 & 1 \\
    Ne-En & 0.363 & 0.365 & 0.263 & 86 & -0.009 & -0.040 & -0.032 & 1 & 0.215 & 0.259 & 0.210 & 1 & 0.511 & 0.491 & 0.374 & 1 \\
    Si-En & 0.284 & 0.283 & 0.203 & 33 & -0.019 & -0.017 & -0.011 & 5 & 0.287 & 0.223 & 0.164 & 5 & 0.462 & 0.477 & 0.351 & 5 \\
    \hline
    \end{tabular}
    }
    \caption{The complete results of the ICL experiment with 7 examples using our proposed AG prompt template (7-ICL-AG). The results include Pearson (r), Spearman ($\rho$), and Kendall's Tau ($\tau$) correlation scores. `LP'-> Language Pair, `E'-> the number of rows excluded because the outputs generated by the large language models did not include a score.}
    \label{tab:updated_icl_ag}
\end{table}

 \clearpage % Ensure the appendix starts on a new page

\section{Appendix: Complete results of unified multilingual training based fine-tuned experiments} \label{app:multi_all}

\begin{table}[ht]
    \centering
    \small
    \resizebox{\textwidth}{!}{%
    \begin{large}  % or \footnotesize, \scriptsize, \tiny
    \begin{tabular}{c|ccc|ccc|ccc|ccc|ccc|ccc}
    \hline
    \multirow{2}{*}{LP} & \multicolumn{3}{c|}{Gemma-7B} & \multicolumn{3}{c|}{Llama-2-7B} & \multicolumn{3}{c|}{Llama-2-13B} & \multicolumn{3}{c|}{OC-3.5-7B} & \multicolumn{3}{c|}{TransQuest} & \multicolumn{3}{c}{CometKiwi} \\
    \cline{2-19}
    & r & $\rho$ & $\tau$ & r & $\rho$ & $\tau$ & r & $\rho$ & $\tau$ & r & $\rho$ & $\tau$ & r & $\rho$ & $\tau$ & r & $\rho$ & $\tau$ \\
    \hline
    
    En-Gu & 0.628 & 0.566 & 0.424 & 0.551 & 0.461 & 0.339 & 0.558 & 0.465 & 0.345 & 0.616 & 0.554 & 0.418 & 0.680 & 0.630 & 0.460 & 0.678 & 0.637 & 0.467 \\
    En-Hi & 0.570 & 0.449 & 0.333 & 0.490 & 0.332 & 0.242 & 0.486 & 0.322 & 0.235 & 0.585 & 0.458 & 0.341 & 0.610 & 0.478 & 0.336 & 0.648 & 0.615 & 0.446 \\
    En-Mr & 0.631 & 0.551 & 0.401 & 0.573 & 0.516 & 0.376 & 0.589 & 0.505 & 0.369 & 0.631 & 0.545 & 0.397 & 0.658 & 0.606 & 0.434 & 0.618 & 0.546 & 0.390 \\
    En-Ta & 0.584 & 0.502 & 0.382 & 0.488 & 0.464 & 0.341 & 0.533 & 0.471 & 0.351 & 0.548 & 0.509 & 0.385 & 0.650 & 0.603 & 0.435 & 0.711 & 0.635 & 0.455 \\
    En-Te & 0.179 & 0.242 & 0.175 & 0.228 & 0.258 & 0.188 & 0.227 & 0.258 & 0.190 & 0.211 & 0.267 & 0.195 & 0.330 & 0.358 & 0.247 & 0.310 & 0.338 & 0.235 \\
    Et-En & 0.688 & 0.728 & 0.534 & 0.594 & 0.636 & 0.455 & 0.622 & 0.655 & 0.469 & 0.643 & 0.678 & 0.493 & 0.755 & 0.760 & 0.560 & 0.853 & 0.860 & 0.661 \\
    Ne-En & 0.688 & 0.650 & 0.476 & 0.598 & 0.519 & 0.370 & 0.628 & 0.565 & 0.404 & 0.657 & 0.607 & 0.438 & 0.767 & 0.718 & 0.530 & 0.783 & 0.789 & 0.599 \\
    Si-En & 0.469 & 0.455 & 0.320 & 0.408 & 0.395 & 0.275 & 0.410 & 0.403 & 0.281 & 0.489 & 0.481 & 0.339 & 0.627 & 0.579 & 0.413 & 0.730 & 0.703 & 0.515 \\
    \hline
    \end{tabular}
    \end{large}
    }
    \caption{The complete results of the UMT instruction fine-tuning experiment with large language models and pre-trained encoder-based approaches (TransQuest-InfoXLM, CometKiwi-XLM-R-XL) for low-resourced language pairs (LP). The results include Pearson (r), Spearman ($\rho$), and Kendall's Tau ($\tau$) correlation scores.}
    \label{tab:comparison}
\end{table}

% \clearpage % Ensure the appendix starts on a new page

\section{Appendix:  Complete results of independent language-pair training based fine-tuned experiments}  \label{app:mono_all}

\begin{table}[ht]
    \centering
    \resizebox{\textwidth}{!}{%
    \begin{large}
    \begin{tabular}{c|ccc|ccc|ccc|ccc|ccc}
    \hline
    \multirow{2}{*}{LP} & \multicolumn{3}{c|}{Gemma-7B} & \multicolumn{3}{c|}{Llama-2-7B} & \multicolumn{3}{c|}{Llama-2-13B} & \multicolumn{3}{c|}{OC-3.5-7B} & \multicolumn{3}{c}{TransQuest} \\
    \cline{2-16}
    & r & $\rho$ & $\tau$ & r & $\rho$ & $\tau$ & r & $\rho$ & $\tau$ & r & $\rho$ & $\tau$ & r & $\rho$ & $\tau$ \\
    \hline
    En-Gu & 0.531 & 0.440 & 0.326 & 0.189 & 0.214 & 0.153 & 0.463 & 0.421 & 0.311 & 0.583 & 0.520 & 0.388 & 0.690 & 0.653 & 0.477 \\
    En-Hi & 0.482 & 0.375 & 0.276 & 0.317 & 0.282 & 0.204 & 0.406 & 0.336 & 0.247 & 0.575 & 0.474 & 0.354 & 0.134 & 0.119 & 0.080 \\
    En-Mr & 0.617 & 0.557 & 0.407 & 0.548 & 0.509 & 0.371 & 0.555 & 0.501 & 0.364 & 0.630 & 0.554 & 0.406 & 0.508 & 0.629 & 0.447 \\
    En-Ta & 0.544 & 0.475 & 0.355 & 0.398 & 0.375 & 0.274 & 0.459 & 0.441 & 0.326 & 0.551 & 0.509 & 0.379 & 0.268 & 0.303 & 0.205 \\
    En-Te & 0.135 & 0.217 & 0.155 & 0.202 & 0.263 & 0.193 & 0.202 & 0.261 & 0.191 & 0.211 & 0.271 & 0.199 & 0.079 & 0.087 & 0.059 \\
    Et-En & 0.622 & 0.648 & 0.467 & 0.569 & 0.589 & 0.417 & 0.559 & 0.598 & 0.421 & 0.609 & 0.652 & 0.470 & 0.797 & 0.806 & 0.603 \\
    Ne-En & 0.660 & 0.612 & 0.441 & 0.545 & 0.497 & 0.352 & 0.582 & 0.543 & 0.388 & 0.646 & 0.614 & 0.444 & 0.777 & 0.746 & 0.554 \\
    Si-En & 0.402 & 0.387 & 0.269 & 0.351 & 0.332 & 0.230 & 0.366 & 0.346 & 0.240 & 0.456 & 0.441 & 0.310 & 0.619 & 0.581 & 0.414 \\
    \hline
    \end{tabular}
    \end{large}
    }
    \caption{The complete results of the ILT instruction fine-tuning experiment with large language models and pre-trained encoder-based approach (TransQuest-InfoXLM) for low-resourced language pairs (LP). The results include Pearson (r), Spearman ($\rho$), and Kendall's Tau ($\tau$) correlation scores.}
    \label{tab:comparison}
\end{table}

\clearpage

\section{Appendix: Examples from error analysis of English-Tamil translation QE task} \label{tab:error_table}

 \begin{figure*}[htbp]
 \centering
 \includegraphics[width=0.85\textwidth, keepaspectratio]{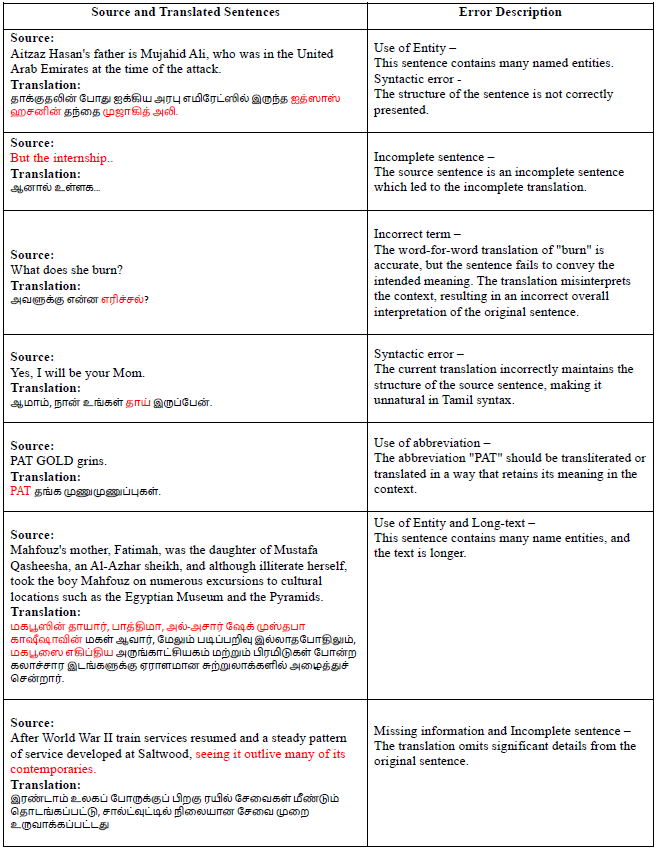}
\caption{ The examples are taken from our study (See in section  \ref{sec:erroranalysis}) analyzing the causes of errors leading to high deviations between human-annotated and predicted DA scores from the best-performing LLM OpenChat for English-Tamil language pair. The words highlighted in red indicate the specific terms causing these errors.
}
%\label{fig:token} % Give a unique label
\end{figure*}
 \clearpage % Ensure the appendix starts on a new page

\section{Appendix: Comparative analysis of results from LLMs in different experimental settings } \label{app:all_setting_spearman}

 \begin{figure}[H]
 \centering
 \includegraphics[width=0.95\textwidth, keepaspectratio]{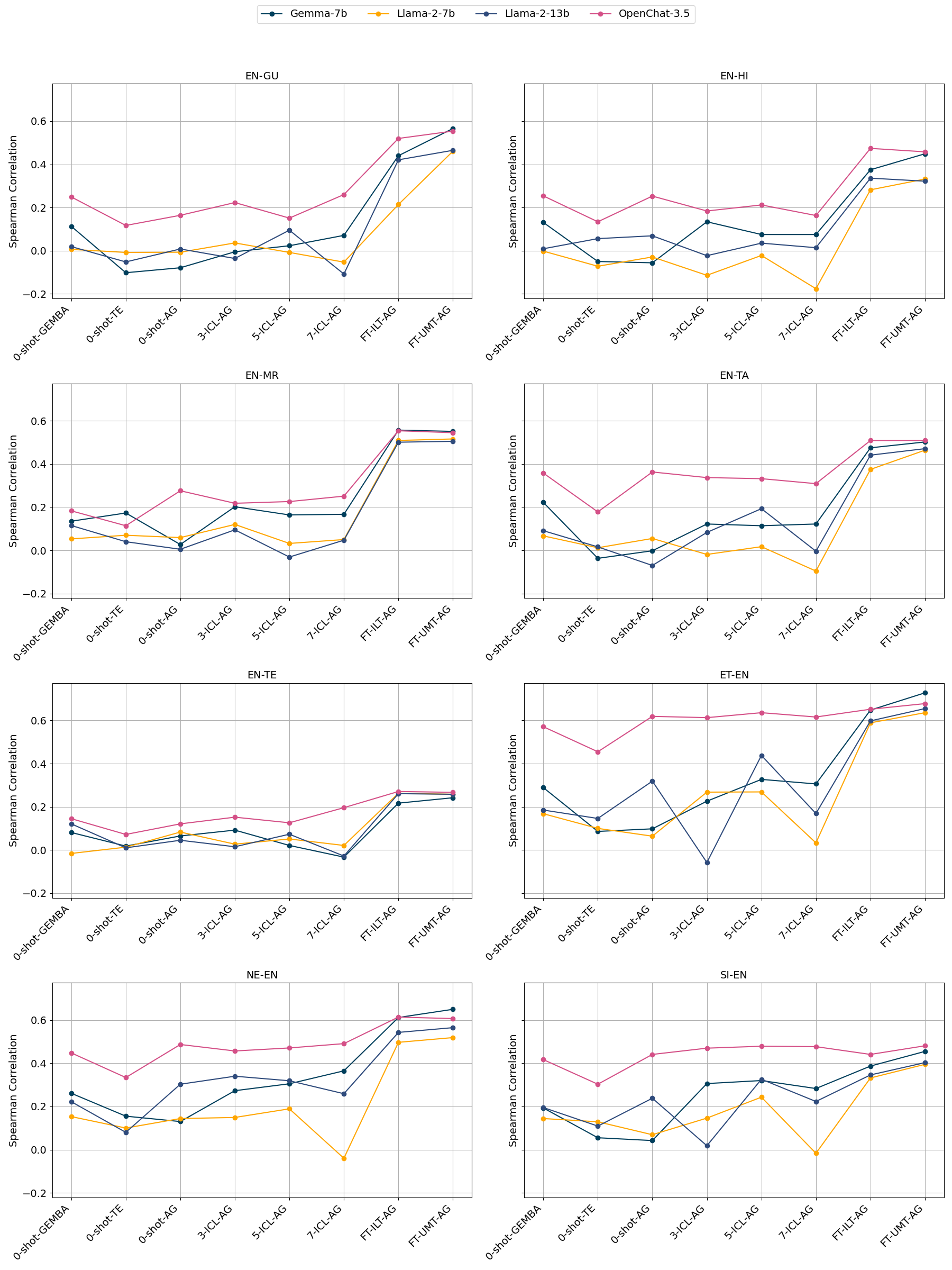}
\caption{ The above graphs show how the Spearman scores varied for each experimental setting with different LLMs. 0-shot-\{ GEMBA, TE, AG \}-> Zero-shot setting with GEMBA, TE and AG prompts;\{N\}-ICL-AG -> In-Context-Learning with N number of examples (N = 3, 5, 7) using AG prompt; FT- \{ILT, UMT\}-AG -> Fine-Tuning with the ILT and UMT setting with the AG prompt.
}
%\label{fig:token} % Give a unique label
\end{figure}
 \clearpage % Ensure the appendix starts on a new page

\section{Appendix: Models, size and disk space utilization} \label{app:model_size}
% \begin{table}[ht!]
% \centering
% \small % Reduces the font size
% \setlength{\tabcolsep}{5pt} % Reduces the space between columns
% \begin{tabular}{@{}lcc@{}}
% \hline
% \toprule
% \\
% \textbf{Model} & \textbf{Model Size (No of Parameters)} & \textbf{Disk Space (GB)} \\ 
% \\
% \hline
% \\
% OpenChat 3.5 & 7 B & 15  \\ \\
% Llama2-7B & 7 B & 27  \\ \\
% Llama2-13B & 13 B & 50  \\ \\
% InfoXLM-large & 560 M & 2.4  \\ \\
% XLM-R-XL & 3.5 B & 13.9  \\ \\
% \bottomrule
% \end{tabular}
% \caption{Model sizes and disk space requirements for various language models used in our study.}
% \label{tab:model_sizes}
% \end{table}

\begin{figure*}[ht]
\captionsetup{size=small}
\centering
\includegraphics[width=0.75\textwidth, keepaspectratio]{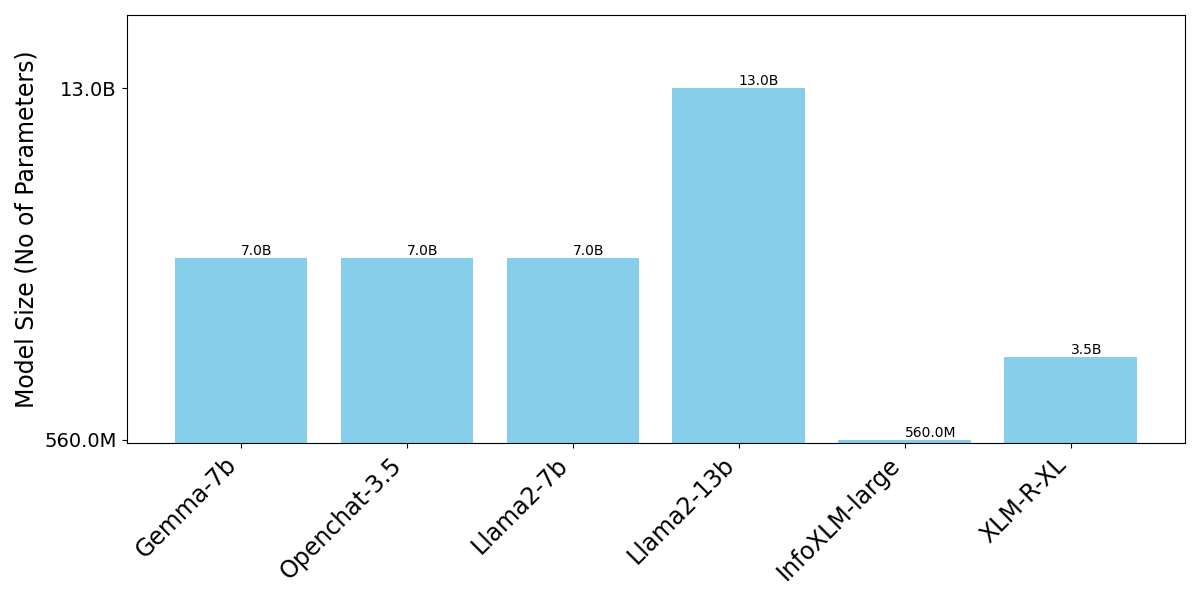}
\caption{This bar graph shows the size (number of parameters) of the large language models we have utilized for our experiments}
%\hfill % Adds horizontal space between the images
\includegraphics[width=0.75\textwidth, keepaspectratio]{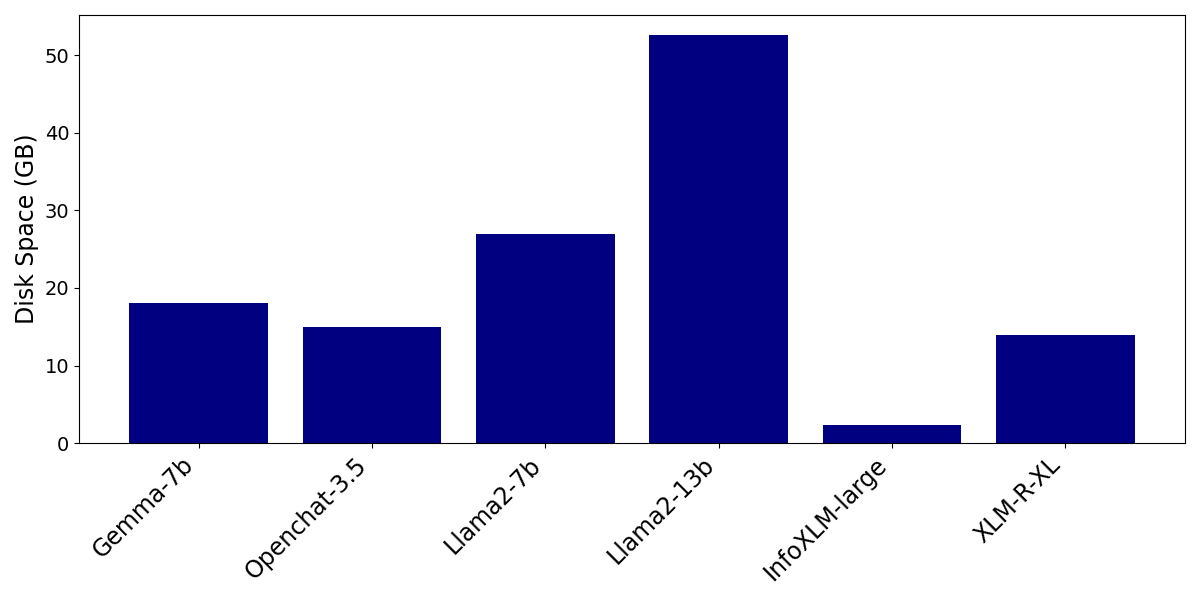}
\caption{This bar graph shows the disk space utilization of the large language models we have utilized for our experiments}
\label{fig:all3modelscatter}
\end{figure*}\label{fig:high_low_resource}

 \clearpage % Ensure the appendix starts on a new page

% \begin{table*}[htbp]
% \centering
% \renewcommand{\arraystretch}{1.5}
% \resizebox{\textwidth}{!}{
% \scriptsize
% \begin{tabular}{|c|c|c|c|c|c|}
% \hline
% \textbf{Lang-pair} & \textbf{Gemma-7B} & \textbf{LLaMA-2-7B} & \textbf{LLaMA-2-13B} & \textbf{OpenChat-3.5} & \textbf{TransQuest} \\
% \hline
% En-Gu & 0.440 & 0.214 & 0.421 & \textcolor{blue}{0.520} & \textcolor{red}{0.6533} \\
% En-Hi & 0.375 & 0.282 & 0.336 & \textcolor{red}{0.474} & 0.1191 \\
% En-Mr & \textcolor{blue}{0.557} & 0.509 & 0.501 & 0.554 & \textcolor{red}{0.6289} \\
% En-Ta & 0.475 & 0.375 & 0.441 & \textcolor{red}{0.509} & 0.3029 \\
% En-Te & 0.217 & 0.263 & 0.261 & \textcolor{red}{0.271} & 0.0866 \\
% Et-En & 0.648 & 0.589 & 0.598 & \textcolor{blue}{0.652} & \textcolor{red}{0.8058} \\
% Ne-En & 0.612 & 0.497 & 0.543 &  \textcolor{blue}{0.614} & \textcolor{red}{0.7456} \\
% Si-En & 0.387 & 0.332 & 0.346 & \textcolor{blue}{0.441} & \textcolor{red}{0.5805} \\
% \hline
% \end{tabular}
% }
% \caption{This table presents the Spearman correlation ($\rho$) scores between the predicted and human-annotated scores for ILT fine-tuning experiments. The scores highlighted in blue represent the highest Spearman scores obtained among the LLMs. The scores highlighted in red indicate the overall best Spearman scores achieved, including both LLMs and pre-trained encoder-based models.}
% \label{tab:ILT_result}
% \end{table*}

\section{Appendix: Our publicly available Hugging Face models} \label{app:trained_models}

\begin{table}[htbp]
\centering
\small
\begin{tabular}{|l|>{\raggedright\arraybackslash}p{10cm}|}
\hline
\textbf{Model} & \textbf{Model Link} \\ \hline
Gemma-7B & \href{https://huggingface.co/ArchSid/AG-Gemma-7B}{ArchSid/AG-Gemma-7B} \\ \hline
Llama-2-7b & \href{https://huggingface.co/ArchSid/AG-Llama-2-7b}{ArchSid/AG-Llama-2-7b} \\ \hline
Llama-2-13b & \href{https://huggingface.co/ArchSid/AG-Llama-2-13b}{ArchSid/AG-Llama-2-13b} \\ \hline
Openchat & \href{https://huggingface.co/ArchSid/AG-openchat}{ArchSid/AG-openchat} \\ \hline
\end{tabular}
\caption{ This table shows the links to our Hugging Face models trained using the Unified Multilingual Training setting.}
\label{table:huggingface-links}
\end{table}

\begin{table}[htbp]
\centering
\small
\begin{tabular}{|l|l|>{\raggedright\arraybackslash}p{10cm}|}
\hline
\textbf{Model} & \textbf{Language-Pair} & \textbf{Model Link} \\ \hline
\multirow{8}{*}{Gemma-7B} & En-Gu & \href{https://huggingface.co/ArchSid/En-Gu_Mono-AG-Gemma-7b}{ArchSid/En-Gu\_Mono-AG-Gemma-7b} \\ \cline{2-3}
                          & En-Hi & \href{https://huggingface.co/ArchSid/En-Hi_Mono-AG-Gemma-7b}{ArchSid/En-Hi\_Mono-AG-Gemma-7b} \\ \cline{2-3}
                          & En-Mr & \href{https://huggingface.co/ArchSid/En-Mr_Mono-AG-Gemma-7b}{ArchSid/En-Mr\_Mono-AG-Gemma-7b} \\ \cline{2-3}
                          & En-Ta & \href{https://huggingface.co/ArchSid/En-Ta_Mono-AG-Gemma-7b}{ArchSid/En-Ta\_Mono-AG-Gemma-7b} \\ \cline{2-3}
                          & En-Te & \href{https://huggingface.co/ArchSid/En-Te_Mono-AG-Gemma-7b}{ArchSid/En-Te\_Mono-AG-Gemma-7b} \\ \cline{2-3}
                          & Et-En & \href{https://huggingface.co/ArchSid/Et-En_Mono-AG-Gemma-7b}{ArchSid/Et-En\_Mono-AG-Gemma-7b} \\ \cline{2-3}
                          & Ne-En & \href{https://huggingface.co/ArchSid/Ne-En_Mono-AG-Gemma-7b}{ArchSid/Ne-En\_Mono-AG-Gemma-7b} \\ \cline{2-3}
                          & Si-En & \href{https://huggingface.co/ArchSid/Si-En_Mono-AG-Gemma-7b}{ArchSid/Si-En\_Mono-AG-Gemma-7b} \\ \hline
\multirow{8}{*}{Llama-2-7b} & En-Gu & \href{https://huggingface.co/ArchSid/En-Gu_Mono-AG-Llama-2-7b}{ArchSid/En-Gu\_Mono-AG-Llama-2-7b} \\ \cline{2-3}
                          & En-Hi & \href{https://huggingface.co/ArchSid/En-Hi_Mono-AG-Llama-2-7b}{ArchSid/En-Hi\_Mono-AG-Llama-2-7b} \\ \cline{2-3}
                          & En-Mr & \href{https://huggingface.co/ArchSid/En-Mr_Mono-AG-Llama-2-7b}{ArchSid/En-Mr\_Mono-AG-Llama-2-7b} \\ \cline{2-3}
                          & En-Ta & \href{https://huggingface.co/ArchSid/En-Ta_Mono-AG-Llama-2-7b}{ArchSid/En-Ta\_Mono-AG-Llama-2-7b} \\ \cline{2-3}
                          & En-Te & \href{https://huggingface.co/ArchSid/En-Te_Mono-AG-Llama-2-7b}{ArchSid/En-Te\_Mono-AG-Llama-2-7b} \\ \cline{2-3}
                          & Et-En & \href{https://huggingface.co/ArchSid/Et-En_Mono-AG-Llama-2-7b}{ArchSid/Et-En\_Mono-AG-Llama-2-7b} \\ \cline{2-3}
                          & Ne-En & \href{https://huggingface.co/ArchSid/Ne-En_Mono-AG-Llama-2-7b}{ArchSid/Ne-En\_Mono-AG-Llama-2-7b} \\ \cline{2-3}
                          & Si-En & \href{https://huggingface.co/ArchSid/Si-En_Mono-AG-Llama-2-7b}{ArchSid/Si-En\_Mono-AG-Llama-2-7b} \\ \hline
\multirow{8}{*}{Llama-2-13b} & En-Gu & \href{https://huggingface.co/ArchSid/En-Gu_Mono-AG-Llama-2-13b}{ArchSid/En-Gu\_Mono-AG-Llama-2-13b} \\ \cline{2-3}
                          & En-Hi & \href{https://huggingface.co/ArchSid/En-Hi_Mono-AG-Llama-2-13b}{ArchSid/En-Hi\_Mono-AG-Llama-2-13b} \\ \cline{2-3}
                          & En-Mr & \href{https://huggingface.co/ArchSid/En-Mr_Mono-AG-Llama-2-13b}{ArchSid/En-Mr\_Mono-AG-Llama-2-13b} \\ \cline{2-3}
                          & En-Ta & \href{https://huggingface.co/ArchSid/En-Ta_Mono-AG-Llama-2-13b}{ArchSid/En-Ta\_Mono-AG-Llama-2-13b} \\ \cline{2-3}
                          & En-Te & \href{https://huggingface.co/ArchSid/En-Te_Mono-AG-Llama-2-13b}{ArchSid/En-Te\_Mono-AG-Llama-2-13b} \\ \cline{2-3}
                          & Et-En & \href{https://huggingface.co/ArchSid/Et-En_Mono-AG-Llama-2-13b}{ArchSid/Et-En\_Mono-AG-Llama-2-13b} \\ \cline{2-3}
                          & Ne-En & \href{https://huggingface.co/ArchSid/Ne-En_Mono-AG-Llama-2-13b}{ArchSid/Ne-En\_Mono-AG-Llama-2-13b} \\ \cline{2-3}
                          & Si-En & \href{https://huggingface.co/ArchSid/Si-En_Mono-AG-Llama-2-13b}{ArchSid/Si-En\_Mono-AG-Llama-2-13b} \\ \hline
\multirow{8}{*}{OpenChat} & En-Gu & \href{https://huggingface.co/ArchSid/En-Gu_Mono-AG-openchat}{ArchSid/En-Gu\_Mono-AG-openchat} \\ \cline{2-3}
                          & En-Hi & \href{https://huggingface.co/ArchSid/En-Hi_Mono-AG-openchat}{ArchSid/En-Hi\_Mono-AG-openchat} \\ \cline{2-3}
                          & En-Mr & \href{https://huggingface.co/ArchSid/En-Mr_Mono-AG-openchat}{ArchSid/En-Mr\_Mono-AG-openchat} \\ \cline{2-3}
                          & En-Ta & \href{https://huggingface.co/ArchSid/En-Ta_Mono-AG-openchat}{ArchSid/En-Ta\_Mono-AG-openchat} \\ \cline{2-3}
                          & En-Te & \href{https://huggingface.co/ArchSid/En-Te_Mono-AG-openchat}{ArchSid/En-Te\_Mono-AG-openchat} \\ \cline{2-3}
                          & Et-En & \href{https://huggingface.co/ArchSid/Et-En_Mono-AG-openchat}{ArchSid/Et-En\_Mono-AG-openchat} \\ \cline{2-3}
                          & Ne-En & \href{https://huggingface.co/ArchSid/Ne-En_Mono-AG-openchat}{ArchSid/Ne-En\_Mono-AG-openchat} \\ \cline{2-3}
                          & Si-En & \href{https://huggingface.co/ArchSid/Si-En_Mono-AG-openchat}{ArchSid/Si-En\_Mono-AG-openchat} \\ \hline
\end{tabular}
\caption{This table shows the links to our Hugging Face models trained using the Independent Language-Pair training setting.}
\label{table:models}
\end{table}

\end{document}